\documentclass[letterpaper,twocolumn,fleqn]{article} 

\usepackage{ist}
\usepackage{graphicx}
\usepackage{url}
\usepackage{amsmath,amssymb,graphicx}
\usepackage{microtype}
\usepackage{booktabs}
\usepackage{multirow}
\usepackage{lineno}

\usepackage{color}
\definecolor{cvprblue}{rgb}{0.21,0.49,0.74}
\usepackage[breaklinks,colorlinks,allcolors=cvprblue]{hyperref}

\usepackage{siunitx}
\usepackage{nicefrac}
\usepackage{xspace}

\usepackage[capitalize]{cleveref}
\crefname{section}{sec.}{secs.}
\Crefname{section}{Sec.}{Secs.}
\crefname{paragraph}{sec.}{secs.}
\Crefname{paragraph}{Sec.}{Secs.}
\crefname{table}{tab.}{tabs.}
\Crefname{table}{Tab.}{Tabs.}
\crefname{figure}{fig.}{figs.}
\Crefname{figure}{Fig.}{Figs.}
\crefname{equation}{eq.}{eqs.}
\Crefname{equation}{Eq.}{Eqs.}

\makeatletter
\DeclareRobustCommand\onedot{\futurelet\@let@token\@onedot}
\def\@onedot{\ifx\@let@token.\else.\null\fi\xspace}
\newcommand{\etal}{et~al\onedot}
\newcommand{\eg}{e.g\onedot}

\newcommand{\baseline}{\textsc{Baseline}\xspace}
\newcommand{\strategyangloss}{\textsc{AngLoss}\xspace}
\newcommand{\strategyaug}{\textsc{Augment}\xspace}
\newcommand{\strategywb}{\textsc{WbTest}\xspace}
\newcommand{\strategywbtrain}{\textsc{WbTrain}\xspace}

\title{Improving the color accuracy of lighting estimation models}

\author{Zitian Zhang$^\dagger$, Joshua Urban Davis$^\ddagger$, Jeanne Phuong Anh Vu$^\ddagger$, Jiangtao Kuang$^\ddagger$, and Jean-Fran\c{c}ois Lalonde$^\dagger$. \\ $^\dagger$Université Laval, Canada $^\ddagger$Meta, United States}

\date{} % date has an empty field.

\begin{document} 

\maketitle 

\thispagestyle{empty} % prevents the first page to be numbered

%%%%%%%%%%%%%%%%%%%%%%%%%%%%%%%%%%
% Abstract
%%%%%%%%%%%%%%%%%%%%%%%%%%%%%%%%%%

%!TEX root = ../main_cic.tex
\begin{abstract}
Advances in high dynamic range (HDR) lighting estimation from a single image have opened new possibilities for augmented reality (AR) applications. Predicting complex lighting environments from a single input image allows for the realistic rendering and compositing of virtual objects. In this work, we investigate the color robustness of such methods---an often overlooked yet critical factor for achieving visual realism. While most evaluations conflate color with other lighting attributes (\eg, intensity, direction), we isolate color as the primary variable of interest. Rather than introducing a new lighting estimation algorithm, we explore whether simple adaptation techniques can enhance the color accuracy of \emph{existing} models. Using a novel HDR dataset featuring diverse lighting colors, we systematically evaluate several adaptation strategies. Our results show that preprocessing the input image with a pre-trained white balance network improves color robustness, outperforming other strategies across all tested scenarios. Notably, this approach requires no retraining of the lighting estimation model. We further validate the generality of this finding by applying the technique to three state-of-the-art lighting estimation methods from recent literature. Our project webpage is available at: \url{https://lvsn.github.io/coloraccuracy}.
\end{abstract}

%!TEX root = ../main_cic.tex
% The first section title should be wrapped inside a \IEEEraisesectionheading as follows.
% \IEEEraisesectionheading{
  \section{Introduction}
  \label{sec:intro}
% }
% The very first letter of the paper is a 2 line initial drop letter
% followed by the rest of the first word in caps.
% \IEEEPARstart{A}{pproaches} 

Approaches for automatically estimating lighting from images have emerged in the past several years and they have enabled new augmented reality (AR) applications, such as the insertion of realistic virtual content in images. This task aims at recovering a representation of the lighting environment in a scene in high dynamic range (HDR) from a single input image. 

Gardner~\etal~\cite{gardner2017learning} have proposed the first deep learning approach for solving this challenging task. Their method has since been improved along several axes: the accuracy of the predicted HDR~\cite{zhan2021emlight}, the generation of highly detailed environment textures~\cite{phongthawee2024diffusionlight}, the speed of inference to obtain real-time performance~\cite{garon2019fast,somanath2021hdr}, the editability of the estimated lighting to allow for artistic control~\cite{weber2022editable,wang2022stylelight,dastjerdi2023everlight}, and predicting spatially-varying lighting representations~\cite{li2020inverse,srinivasan2020lighthouse,zhu2022irisformer}.

\begin{figure}[t]
\setlength{\tabcolsep}{1pt}
\begin{tabular}{ccc}

\includegraphics[width=0.325\linewidth]{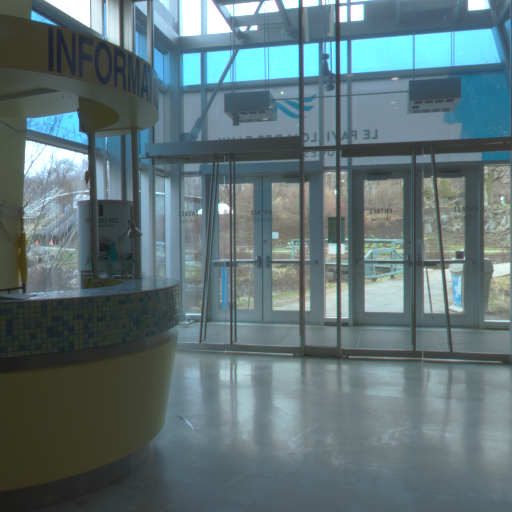} &
\includegraphics[width=0.325\linewidth]{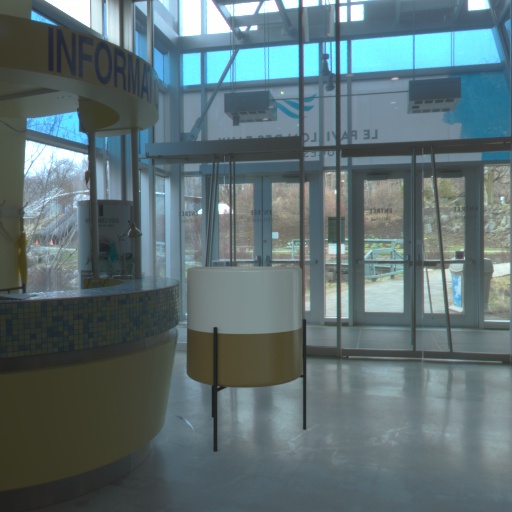} & 
\includegraphics[width=0.325\linewidth]{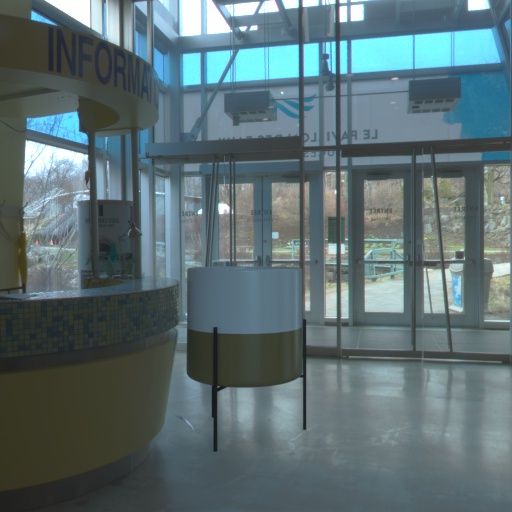} \\

\includegraphics[width=0.325\linewidth]{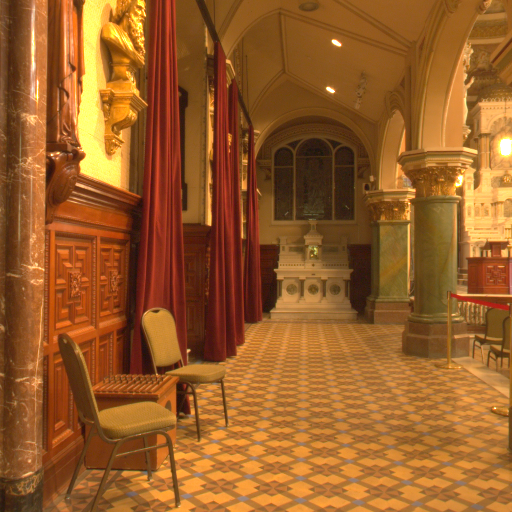} &
\includegraphics[width=0.325\linewidth]{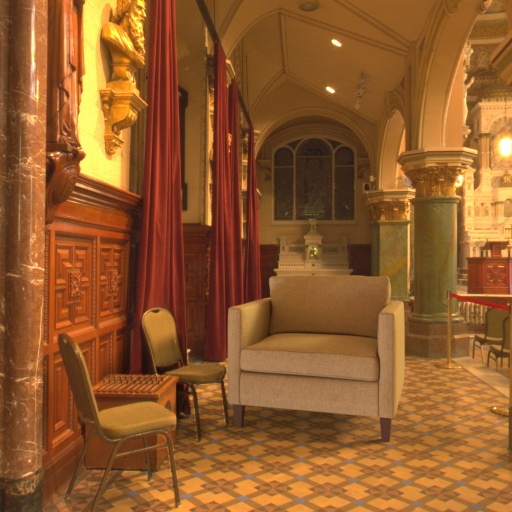} & 
\includegraphics[width=0.325\linewidth]{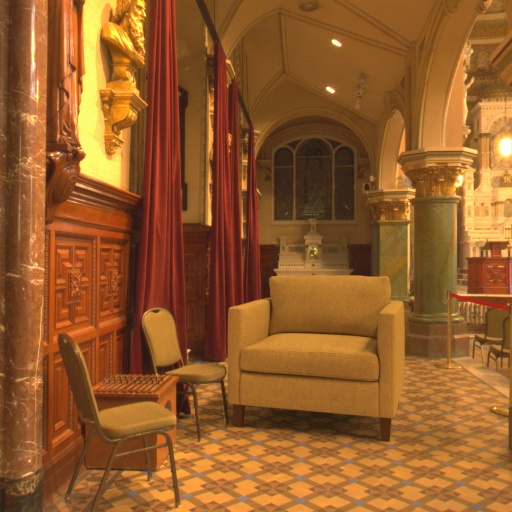} \\

\end{tabular}
% \zitian{Generate examples (maybe more than 1?) of object inserted in a background image where the lighting is correct but color is wrong. Then show same example where color has been fixed w/ \strategywb.}
\caption{Getting the ``color right'' is critical for realistic virtual object insertion. Left: input image. Middle: virtual object (sofa) relit with estimated lighting from a baseline model. Right: a lighting estimation model with higher color accuracy results in a more realistic composite.}
\label{fig:teaser}
\end{figure}

In this literature, not much attention has been given to the color accuracy of such models. Indeed, while methods evaluate themselves on standard benchmarks~\cite{gardner2017learning} and report a variety of quantitative metrics, color accuracy is not specifically targeted. However, accurate color reproduction is a very important factor in realism when inserting virtual objects in images~\cite{reinhard2001color}. Consider the examples in \cref{fig:teaser}, where virtual objects are relit by the prediction of an automatic lighting estimation algorithm. While the overall lighting direction is accurately rendered, the wrong color (\cref{fig:teaser}, middle) makes the object seem out of place. Achieving accurate color rendering by adjusting the lighting estimation model with one of the strategies we explore in this paper (\cref{fig:teaser}, right) results in a much more believable composition. 
%In particular, a wide range of different illuminant are present in images, especially in indoor scenes. They may even contain multiple illuminants of different spectra, making the task even harder. 

In this paper, we aim to systematically study how lighting estimation models produce color-consistent results for a given input image. Specifically, we focus on the case of indoor scenes. To this end, we capture a new test dataset of different indoor environments, where each scene is captured under several different white balance conditions. This simulates the presence of different illuminants in the scenes and, in each case, the corresponding ground truth HDR lighting map is available. We then leverage this dataset, used only at test time, to quantify the impact of various color adaptation strategies on the robustness of the lighting estimation model, especially for the case of virtual object insertion, where an object is relit and composited in the image. 

Our goal is to identify strategies that are agnostic to the specific lighting estimation approach, with the aim of providing practitioners a set of ``best color practices'' they can employ with their own approach. Specifically, we explore the use of color-specific loss function, chromatic data augmentation, and leveraging pre-trained white balance networks at both training and test time. Our experiments show that correcting for the overall color of the input image, with a pre-trained white balance network, achieves the best performance across all tested strategies. This is particularly useful since it requires no re-training of the lighting estimation network itself: it can be used as is. We demonstrate the practical impact of this solution by evaluating the same strategy on two other lighting estimation approaches from the recent literature, and show similar improvements in color accuracy. We hope this paper will both bring attention to the problem of color accuracy in lighting estimation, and provide practitioners with the tools to improve this accuracy in their approaches.

%!TEX root = ../main_cic.tex
\section{Related work}
\label{sec:relwork}

Estimating lighting from images has been explored from a variety of angles in recent years. Here, we focus on works who aim to estimate lighting from a single image captured indoors. We refer the reader to \cite{einabadi2021deep} for a broader overview of deep learning models for lighting estimation. 
Gardner~\etal~\cite{gardner2017learning} pioneered the use of deep learning for lighting estimation in indoor scenes, and proposed a network that can predict an environment map ($360^\circ$ lighting map) in HDR from a single input image. 
% In doing so, they also introduced a large dataset HDR environment maps, dubbed the Laval Indoor HDR Dataset, which has since become the benchmark in lighting estimation evaluation. 
This technique was later improved in several ways. Legendre~\etal~\cite{legendre2019deeplight} train a single model for indoor and outdoor images, Song and Funkhouser~\cite{song2019neural} take into account the object insertion position. Garon~\etal~\cite{garon2019fast} focus on speed and infer fast spatially-varying light estimates using spherical harmonics. %, which have also been used in other context for AR applications \cite{mandl2017learning,zhao2020pointar}. 
% Cheng~\etal~\cite{cheng2018shlight} make the interesting observation that on-device applications can benefit from front and back cameras readily available on mobile devices. 
Somanath and Kurz~\cite{somanath2021hdr} train an UNet and focus on real-time on-device estimation. 
A variety of other lighting parametrizations have been used. For instance, parametric lights~\cite{gardner2019deep} and spherical gaussians~\cite{zhan2021emlight,zhan2021gmlight} typically provide more controllability on the output since they possess few, intuitive parameters. This has explicitly been exploited by \cite{weber2022editable,dastjerdi2023everlight} who combine a parametric lighting with powerful GANs for editable lighting estimation. StyleLight~\cite{wang2022stylelight} employ a GAN-inversion approach to achieve some level of control over the output. It has also been proposed to learn volumetric lighting representations from images. Of note, Lighthouse~\cite{srinivasan2020lighthouse} learns multi-scale volumetric lighting from a stereo pair, \cite{li2020inverse,zhu2022irisformer} predicts a dense 2D grid of spherical gaussians which is further extended into a 3D volumetric representation by Wang~\etal~\cite{wang2021learning}. 
More recently, diffusion models have also been used~\cite{phongthawee2024diffusionlight} for environment map generation. 
% However, most recent work have tried to leverage these models to achieve image relighting without relying on the explicit lighting estimation step. This has been demonstrated on real indoor~\cite{poirier2024diffusion} or outdoor images~\cite{kocsis2024lightit}, generated objects~\cite{sharma2024alchemist,zeng2024dilightnet}, or virtual objects~\cite{zhang2025zerocomp,zeng2024rgb}.
%
Most lighting estimation methods are evaluated on the Laval Indoor HDR Dataset~\cite{gardner2017learning}. To this end, the typical evaluation methodology employed is to render a simple virtual object (e.g., one or many spheres with different material properties~\cite{zhan2021emlight,weber2022editable,wang2022stylelight}) with both the ground truth and estimated lighting, and compute image quality metrics to establish the degree of similarity. While color metrics are commonly used (e.g., RGB angular error~\cite{legendre2019deeplight,wang2022stylelight,dastjerdi2023everlight}), no paper has, to our knowledge, specifically evaluated the color robustness of lighting estimation models.

%!TEX root = ../main_cic.tex
\section{Strategies for improved color accuracy}
\label{sec:method}

We wish to identify methods for improving the color accuracy of lighting estimation models, but that are independent of the specific approach used. In that sense, we avoid approach-specific mechanisms, for example: a specific type of convolutions can only be applicable to CNN-based approaches, and would not work on transformer-based architectures. 

Therefore, we seek to adjust the behavior of a lighting estimation model $\mathcal{L}(I) = L$, which produces a estimated lighting $L$ from a single input image $I$, without modifying $\mathcal{L}$ itself. We also assume the estimated lighting $L$ and its ground truth target $L^*$ can both be expressed as an image (for example, in spherical coordinates through equirectangular projection), although the following could easily be adapted to other representations (e.g., spherical harmonics or spherical gaussians can readily be converted to spherical images). 

% Below we survey different options studied in this work. 

\subsection{Adding a color-specific loss}

The first strategy is to add a color-specific loss, to enforce that the predicted lighting $L$ yields a color as close as possible to the ground truth target $L^*$. We denote
\begin{equation}
\ell_{\mathrm{ang}}
= \frac{1}{4\pi}\sum_{i=1}^{N}\bigl(1 - \cos\!\big(\tilde{L_i},\,\tilde{L_i}^*\big)\bigr)\,d\omega_i,
\label{eq:angloss}
\end{equation}

where $N$ is the total number of pixels in $L$, $i$ is a pixel index in $L$, and $d\omega_i$ the solid angle subtended by that pixel, each $\tilde{L}_i = \tfrac{(R_i,\,G_i,\,B_i)}{R_i+G_i+B_i}$ 
is the chromaticity-normalized RGB vector of pixel $i$ (similarly for $\tilde{L}_i^*$). This loss penalizes large angular differences between the predicted lighting $L$ and the target $L^*$. We name this strategy \strategyangloss.  

% \subsection{Adding a render loss}

\subsection{Chromatic data augmentation}

\begin{figure}
\centering
\footnotesize
\setlength{\tabcolsep}{1pt}
\newlength{\mywidth}
\setlength{\mywidth}{0.19\linewidth}
\begin{tabular}{ccccc}
\includegraphics[width=\mywidth]{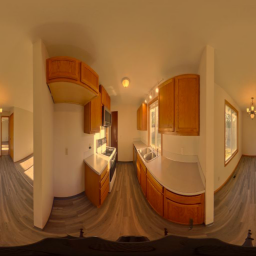} &
\includegraphics[width=\mywidth]{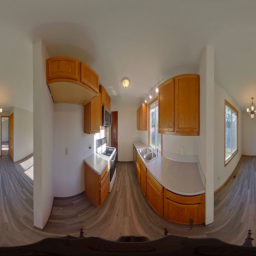} &
\includegraphics[width=\mywidth]{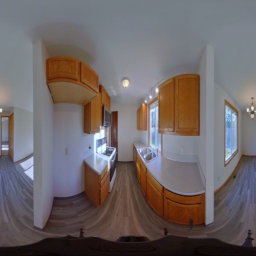} &
\includegraphics[width=\mywidth]{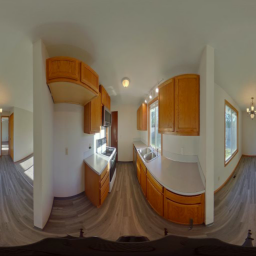} &
\includegraphics[width=\mywidth]{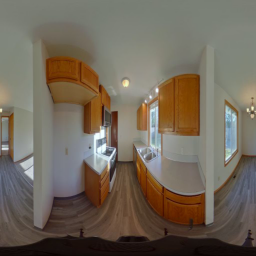} \\

A & B & C & D50 & D55 \\

\includegraphics[width=\mywidth]{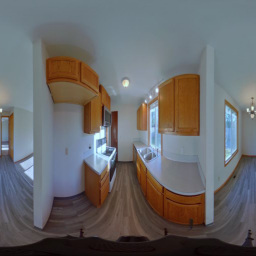} &
\includegraphics[width=\mywidth]{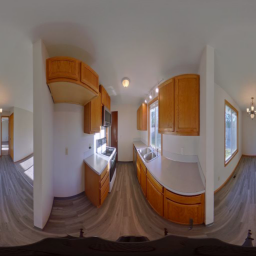} &
\includegraphics[width=\mywidth]{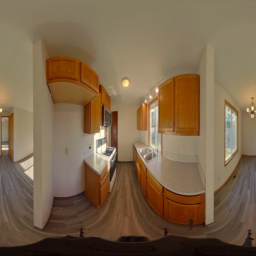} &
\includegraphics[width=\mywidth]{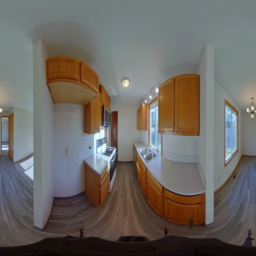} &
\includegraphics[width=\mywidth]{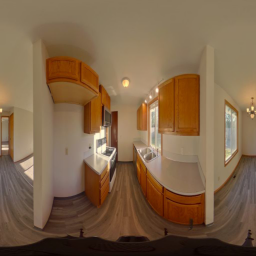} \\

D75 & E & F2 & F7 & F11 \\

\end{tabular}
\caption{Through chromatic adaptation, we augment the training data of the lighting estimation network $\mathcal{L}$ to present it with a wider variety of illuminant during training. The illuminant name is shown below each image.}
\vspace{-1em}
\label{fig:chromatic-adapt}
\end{figure}

Another strategy is to rely on data augmentation and present the network with a wider variety of illuminant color during training. For this, we apply chromatic adaptation to map the target lighting $L^*$ to a different standard illuminant, which are chosen uniformly at random from: (no change), A, B, C, D50, D55, D65, D75, E, F2, F7 and F11. \Cref{fig:chromatic-adapt} shows an example of such data augmentation strategy. Note that, while typical white balance algorithms in real-world ISPs may never create such extreme illuminants, we employ this strategy strictly as a way to provide a greater variability in the training data~\cite{shorten2019survey}. We name this strategy \strategyaug. 

\subsection{Adjusting for white balance}

A last strategy is to leverage a white balance editing network~\cite{afifi2020deepWB} $\mathcal{C}$, which has been trained to convert an input image $I$ to a neutral color version $I^\prime$, i.e., $I^\prime = \mathcal{C}(I)$. We explore its use first at test time only, then during training as well. 

\begin{figure}
\centering
\footnotesize
\includegraphics[width=\linewidth]{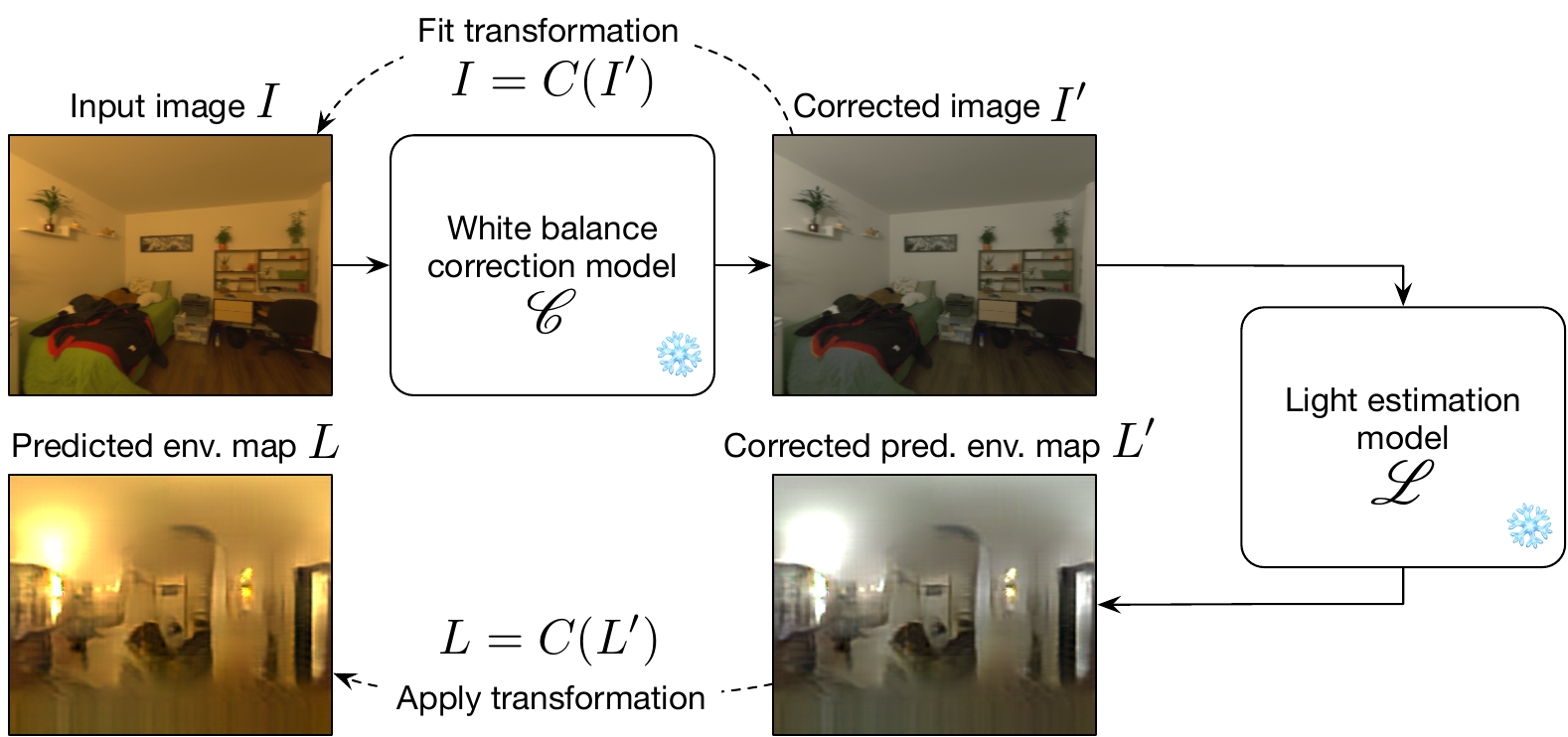} \\
(a) Exploiting a white balance network $\mathcal{C}$ at test time. \\*[0.5em]
\includegraphics[width=\linewidth]{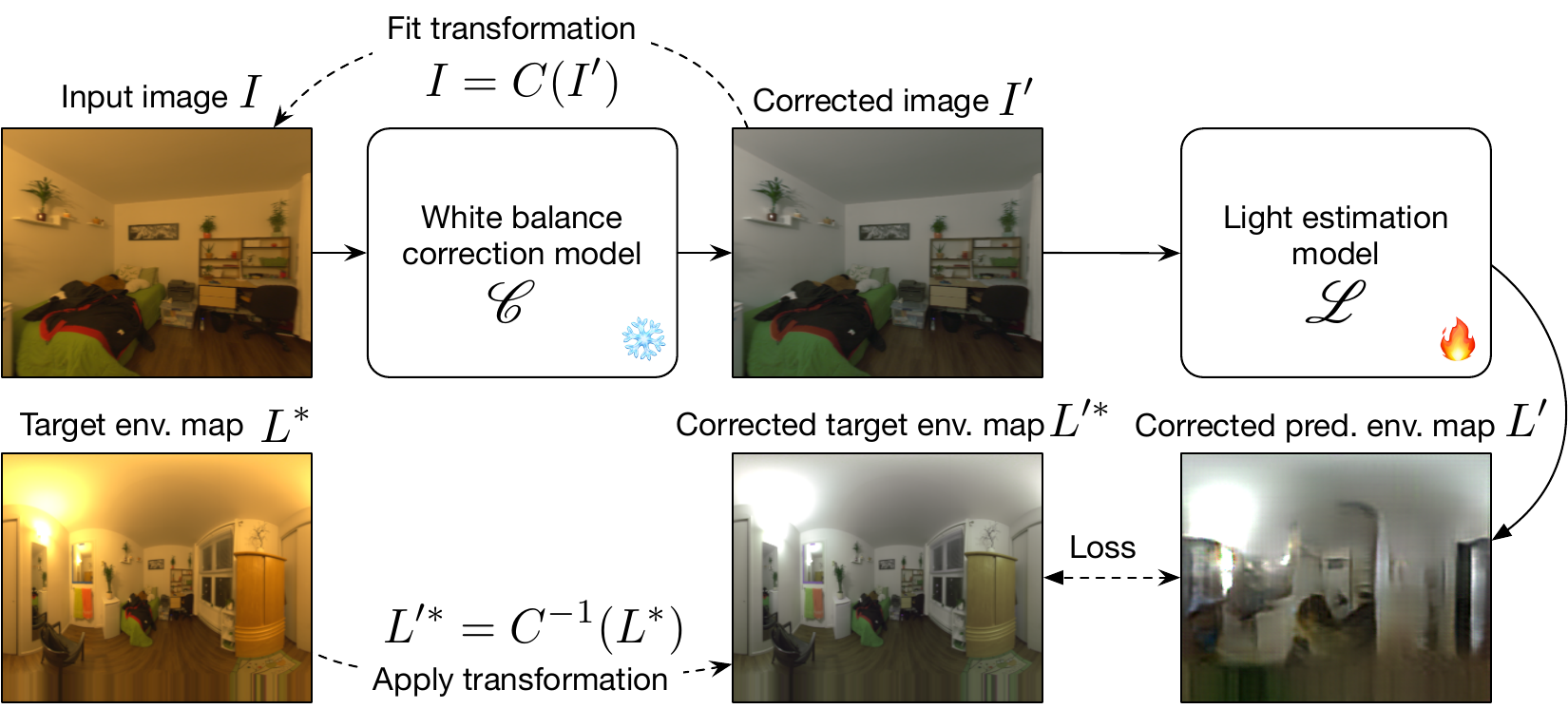} \\
(b) Exploiting a white balance network $\mathcal{C}$ during training. 
\caption{Using a white balance network $\mathcal{C}$ for improving color robustness. We experiment with employing it at (a) test time only, where all networks ($\mathcal{C}$ and $\mathcal{L}$ have frozen weights); and (b) during the training of the lighting estimation network $\mathcal{L}$ as well (only $\mathcal{C}$ is frozen). We posit that removing the burden of reasoning about color from the lighting estimation network $\mathcal{L}$ will make its task easier and thus achieve greater performance.}
\vspace{-1.5em}
\label{fig:white-balance}
\end{figure}

\paragraph{At test time} 
% \label{subsubsec:wb-test}
First, the input image $I$ is color-corrected to obtain $I^\prime$, which is then fed to the lighting estimation model $\mathcal{L}$. The predicted lighting $L^\prime = \mathcal{L}(I^\prime)$ must be corrected back to the original white balance setting of the input by applying the inverse transformation $\mathcal{C}^{-1}$. In practice, $\mathcal{C}$ is a feed-forward neural network, so we cannot assume it to be invertible. We obtain an approximation $C \approx \mathcal{C}^{-1}$ by fitting a simple linear operator (e.g., a $3 \times 3$ color transformation matrix) such that $C(I^\prime) = I$, via a least-squares fit.  The resulting $C$ can then be applied to $L^\prime$ to obtain $L = C(L^\prime)$. Compactly, this strategy amounts to 
\begin{equation}
L = C(L') = C\big(\mathcal{L}(\mathcal{C}(I))\big),
\label{eq:strategywb}
\end{equation}
We dub this strategy \strategywb and \cref{fig:white-balance}-(a) illustrates this process graphically.

\paragraph{During training}
% \label{subsubsec:wb-training}
% 
A similar strategy can also be applied during training. Here, the idea is to train the lighting estimator model on white-balanced images $I^\prime = \mathcal{C}(I)$, with the corresponding ground truth (target) lighting also color-corrected, $\mathcal{C}(L^*)$. To ensure both the input image and target lighting are color-corrected using the same transformation $T$, the same strategy as in \cref{eq:strategywb} is applied to obtain $T$. Then, the corrected target environment map is obtained by $L^{\prime *} = T^{-1}(L^*)$. This is illustrated graphically in \cref{fig:white-balance}. At inference time, \cref{eq:strategywb} is also applied. This focuses training solely on the estimating lighting instead of color. Intuitively, this should result in an easier training since the lighting estimation network $\mathcal{L}$ no longer needs to adapt to the color of the input image. We name this strategy \strategywbtrain.

%!TEX root = ../main_cic.tex

\begin{figure}[t]
\centering
\setlength{\tabcolsep}{2pt}
\newlength{\tmplength}
\setlength{\tmplength}{0.289\linewidth}
% \begin{tabular}{cc}
\includegraphics[width=0.9\linewidth]{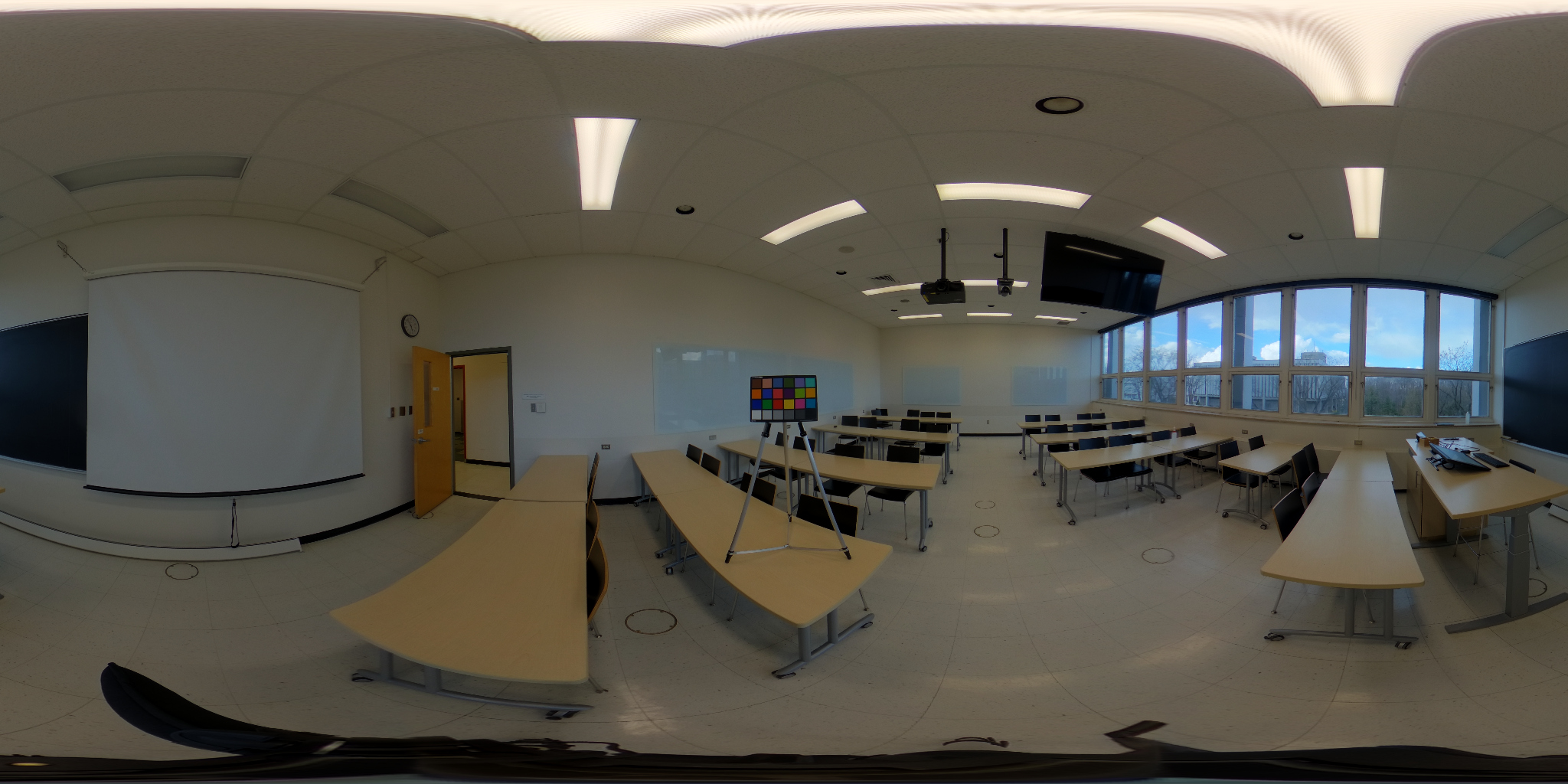} \\
\begin{tabular}{ccc}
\includegraphics[width=\tmplength]{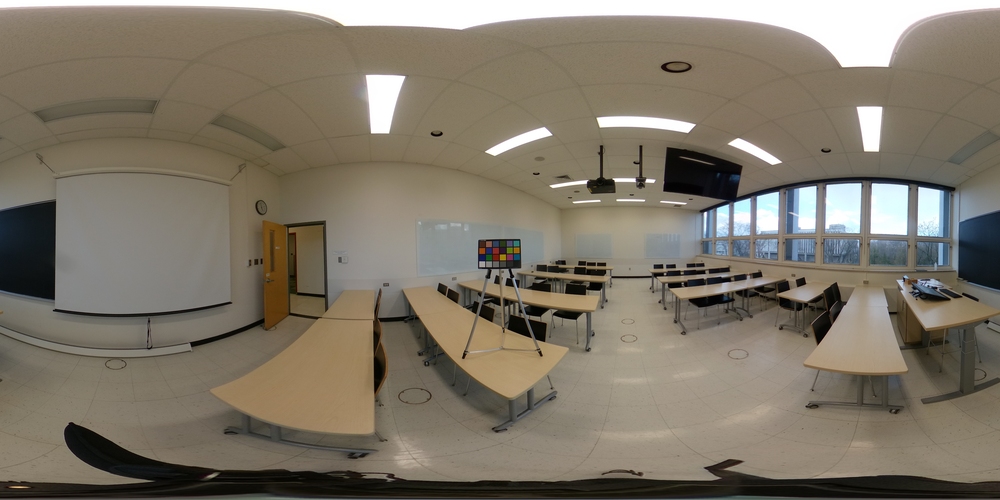} &
\includegraphics[width=\tmplength]{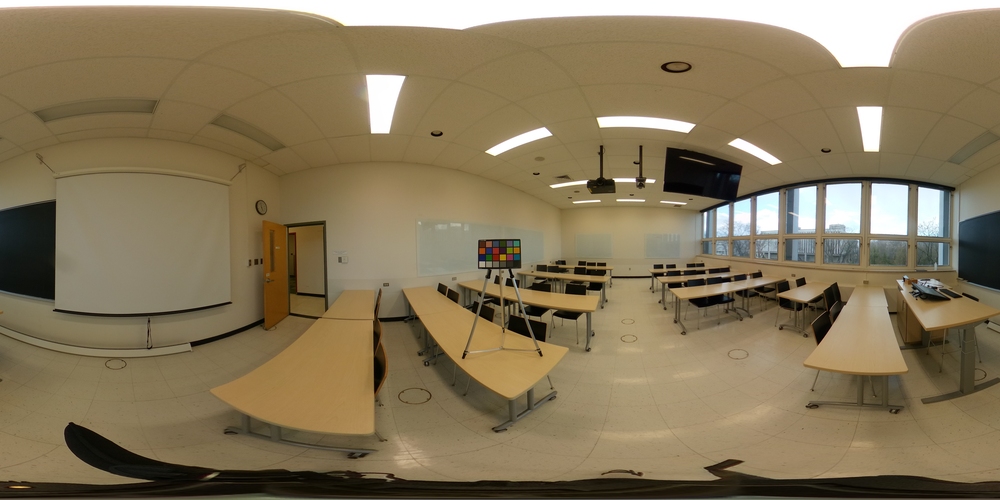} &
\includegraphics[width=\tmplength]{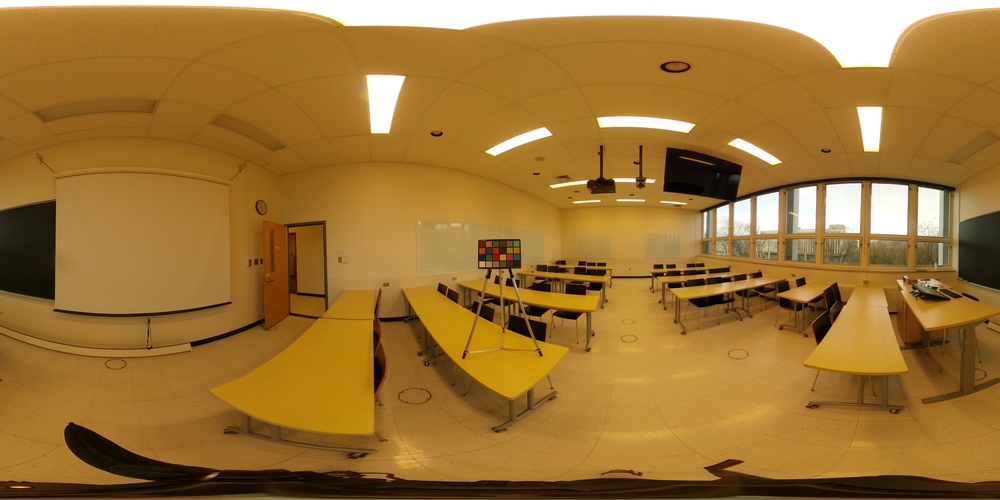} \\
\includegraphics[width=\tmplength]{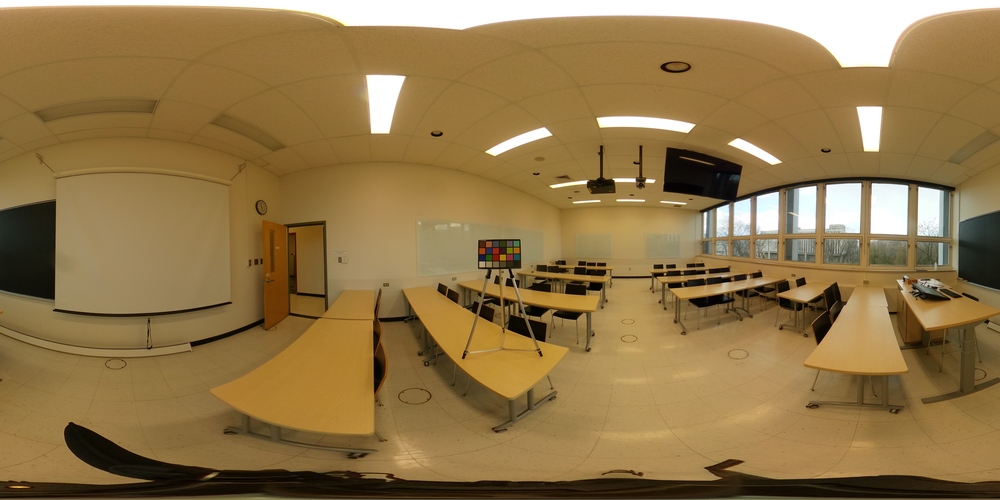} &
\includegraphics[width=\tmplength]{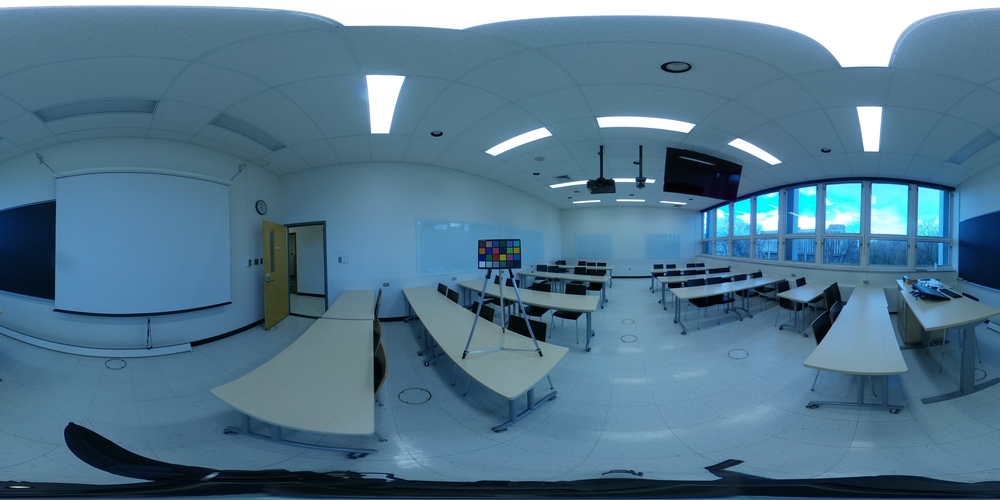} &
\includegraphics[width=\tmplength]{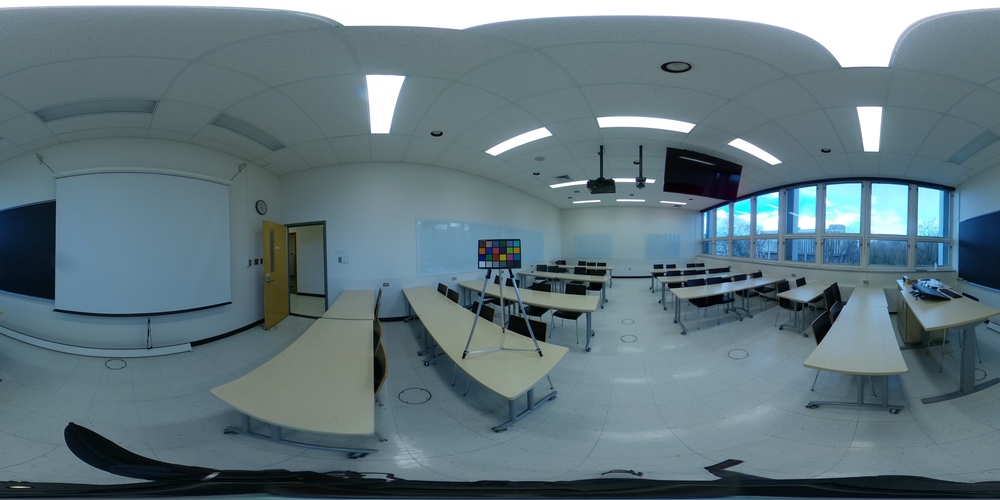} \\
\includegraphics[width=\tmplength]{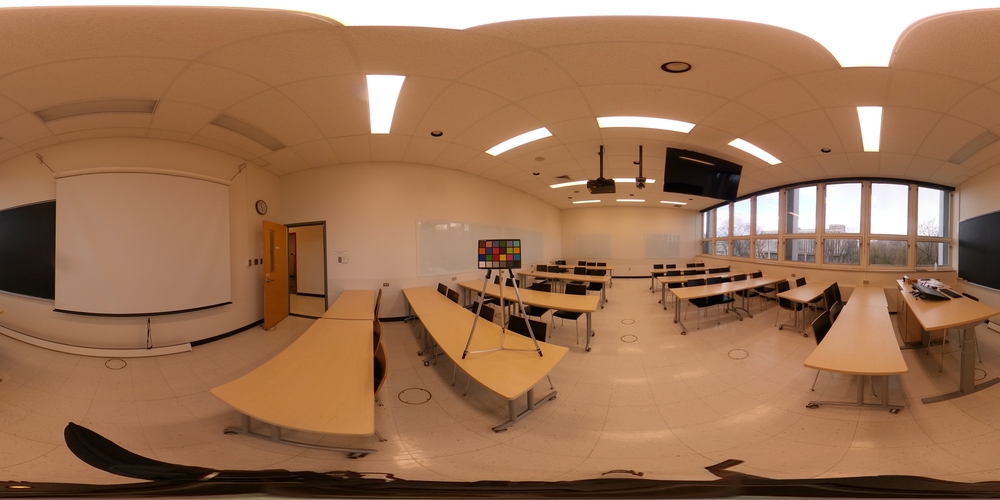} &
\includegraphics[width=\tmplength]{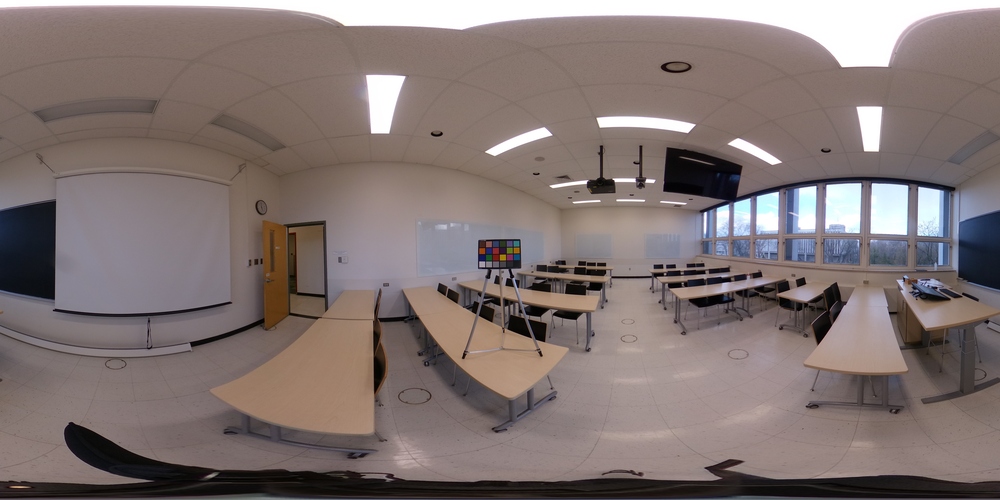} &
\includegraphics[width=\tmplength]{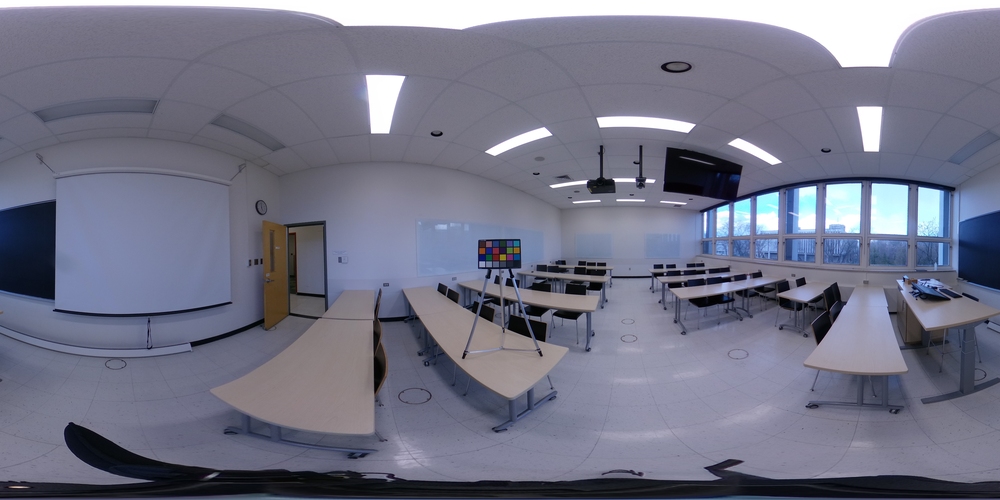} \\
\end{tabular}
% & 
% \begin{tabular}{ccc}
% \includegraphics[width=\tmplength]{figures/dataset/examples/vnd3518/R0016425_AutoWB.JPG} &
% \includegraphics[width=\tmplength]{figures/dataset/examples/vnd3518/R0016426_Daylight.JPG} &
% \includegraphics[width=\tmplength]{figures/dataset/examples/vnd3518/R0016427_Shade.JPG} \\
% \includegraphics[width=\tmplength]{figures/dataset/examples/vnd3518/R0016428_Cloudy.JPG} &
% \includegraphics[width=\tmplength]{figures/dataset/examples/vnd3518/R0016430_Incandescent1.JPG} &
% \includegraphics[width=\tmplength]{figures/dataset/examples/vnd3518/R0016431_Incandescent2.JPG} \\
% \includegraphics[width=\tmplength]{figures/dataset/examples/vnd3518/R0016432_DaylightFluorescent.JPG} &
% \includegraphics[width=\tmplength]{figures/dataset/examples/vnd3518/R0016433_NeutralWhiteFluorescent.JPG} &
% \includegraphics[width=\tmplength]{figures/dataset/examples/vnd3518/R0016436_CoolWhiteFluorescent.JPG} \\
% \end{tabular}
% \end{tabular}
\caption{Capturing scenes under different white balance settings. For two different indoor scenes captured in HDR (top, tone-mapped with \cite{reinhard2005dynamic} for display), we also capture a series of JPG images under different white balance settings (bottom), to acquire a rich dataset of realistic color renderings. A color checker is placed in each scene for accurate color calibration.}
\vspace{-1.5em}
\label{fig:capture-protocol}
\end{figure}

\begin{figure}
\centering
\footnotesize
\setlength{\tabcolsep}{1pt}
\setlength{\tmplength}{0.24\linewidth}
\begin{tabular}{ccccc}
\includegraphics[width=\tmplength]{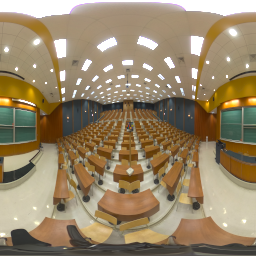} &
\includegraphics[width=\tmplength]{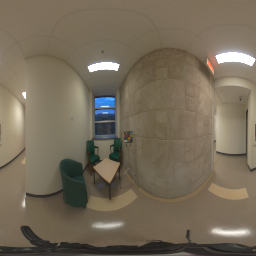} &
\includegraphics[width=\tmplength]{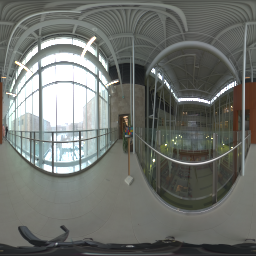} &
\includegraphics[width=\tmplength]{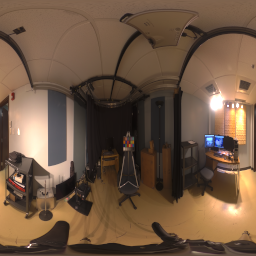} \\

\includegraphics[width=\tmplength]{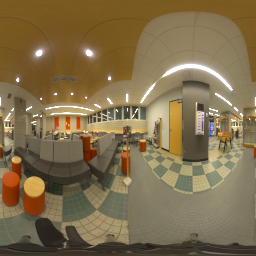} &
\includegraphics[width=\tmplength]{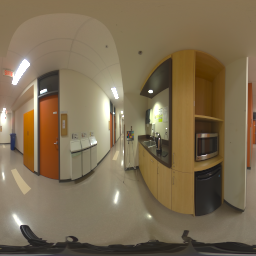} &
\includegraphics[width=\tmplength]{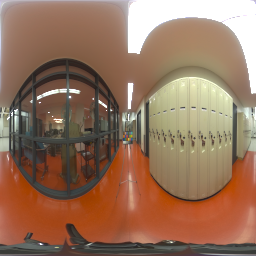} &
\includegraphics[width=\tmplength]{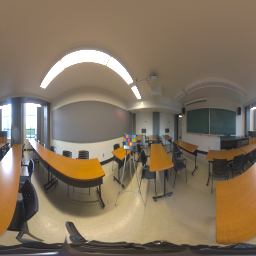} \\

\includegraphics[width=\tmplength]{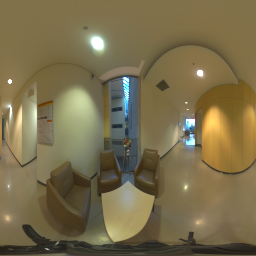} &
\includegraphics[width=\tmplength]{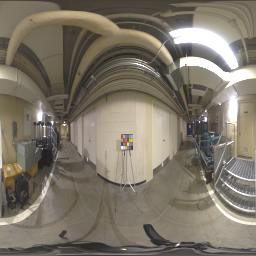} &
\includegraphics[width=\tmplength]{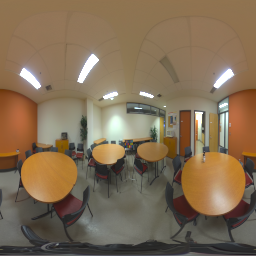} &
\includegraphics[width=\tmplength]{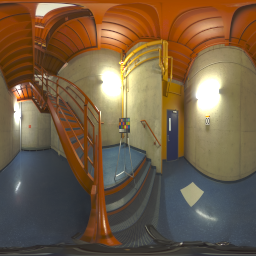} \\

\includegraphics[width=\tmplength]{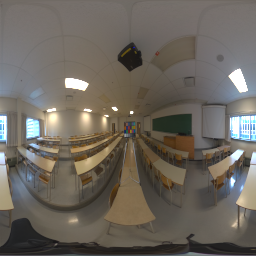} &
\includegraphics[width=\tmplength]{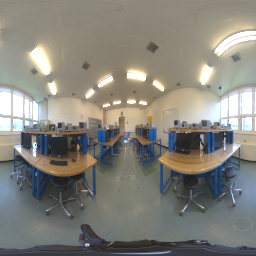} &
\includegraphics[width=\tmplength]{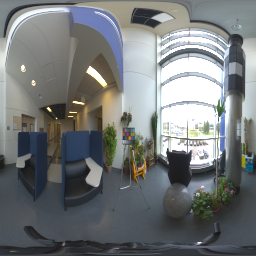} &
\includegraphics[width=\tmplength]{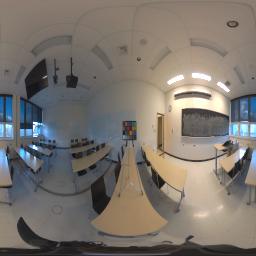} \\

\end{tabular}
\caption{Example scenes (tonemapped for display) in the test dataset. 
}
\label{fig:dataset-thumb}
\end{figure}

\section{Dataset}
\label{sec:dataset}

While there exists a variety of HDR lighting datasets~\cite{gardner2017learning,murmann2019dataset,kim2021large}, none are particularly appropriate to evaluate the color accuracy of lighting estimation methods as they either focus on dynamic range~\cite{gardner2017learning}, lighting directions~\cite{murmann2019dataset}, or do not contain full $360^\circ$ lighting~\cite{kim2021large}. Therefore, we captured a novel dataset of HDR $360^\circ$ panoramas in a variety of indoor scenes along with in-camera JPG images under a variety of different white balance settings. This dataset, specifically targeted to the task of evaluating color robustness, accurately replicates a real-world scenario that would be hard to reproduce using data augmentation because we capture images with the hardware ISP on-board the camera. 

\subsection{Capture protocol}

For a given scene, we employ the following capture protocol. First, a Ricoh Theta Z1 camera is placed on a tripod in (approximately) the center of the scene. A Macbeth color chart is also installed on a small tripod and placed approximately $1.5$m in front of the camera, in an area of mostly uniform lighting. %, and not in front of a bright light source (e.g., a window) to make sure it gets properly imaged by the Z1 camera. 

As in \cite{bolduc2023beyond}, a set of 11 exposures are captured in exposure bracketing mode, with the camera set in RAW. The resulting files are subsequently merged into a single EXR file using \cite{debevec1997recovering}. After HDR capture, a photo is captured with the Theta Z1 camera in all white balance settings supported by the camera, namely: auto, daylight, shade, cloudy, incandescent 1, incandescent 2, daylight fluorescent, neutral white fluorescent, cool white fluorescent, and warm white fluorescent. This is, of course, not equivalent to physically changing the properties of the illuminants, as was done in the LSMI dataset~\cite{kim2021large} by turning on/off individual light sources. Unfortunately, the HDR environment maps were not captured so full lighting is not available. In our case, we approximate the change of illuminant in a global fashion through the white balance setting onboard the camera ISP. The color checker (present in all frames) is used to adjust the white balance of the HDR panorama to each of the white balance settings. 

% \paragraph{Handheld camera} Finally, a handheld camera (we used a Canon PowerShot A3500 IS) is used to capture between 1 and 5 shots in a few directions around the scene. At least 1 shot is always aimed parallel to the front view of the Z1 camera, in order to have a common view with the Macbeth chart. The camera is set in auto-exposure mode with the integrated flash function disabled when capturing these images. 

\subsection{Dataset overview}

Example images for two scenes from our dataset are shown in \cref{fig:capture-protocol}. The variety of white balance settings simulates vastly different illuminants. \Cref{fig:dataset-thumb} shows a preview of additional scenes in our dataset. As can be observed in the figure, a wide variety of indoor lighting conditions are present in the dataset, either coming from different scenes (e.g., hallways, classrooms, meeting rooms, kitchens, corridors, basements, parking lots, etc.), or from the same scene but with different lighting conditions. 

In all, a total of 2,233 images over 78 different scenes are captured. For each panorama, we extract three perspective images at $90^\circ$ field of view, at $120^\circ$ intervals, to be used as input images to the lighting estimation networks, at test time only. The dataset is available publicly at the project webpage.

\section{Evaluation}
\label{sec:evaluation}

We now proceed to evaluate the proposed strategies on our new test dataset. Note that the dataset is used only at test time: none of its images are used during training. 

\subsection{Baseline lighting estimation method}
\label{sec:baseline}

We define a simple \baseline lighting estimation method on which to test potential color adaptation strategies. 

\paragraph{Network architecture}
We draw inspiration from \cite{somanath2021hdr,weber2022editable} and adopt the conditional GAN architecture of pix2pixHD~\cite{wang2018high}. This network is composed of a CNN-based generator and discriminator. The generator follows an encoder-decoder setup in which the input image is downsampled using a series of strided convolutional layers followed by a bottleneck of residual blocks, the output of which is then upsampled to the original resolution using a series of transpose convolutions. The multi-scale discriminator~\cite{wang2018high} contains two seperate networks: the first operates at the original image resolution and receives a concatenation of the input image and the generator prediction as input, whereas the second operates at half the scale and takes a (average-pooled) downsampled version of the aforementioned concatenation as input.

The network takes as input a single (square) LDR image $I \in \mathbb{R}^{256 \times 256 \times 3}$, and directly outputs a full \ang{360} HDR panorama $L$ (equirectangular projection) of the same spatial resolution. To help the network deal with high dynamic range targets, the network outputs are passed through a (differentiable) inverse tonemapping function $f_\mathrm{iTM}$ prior to computing the losses, $f_\mathrm{iTM}(x) = x^{2.2}$.

\paragraph{Training details} 
The Laval Indoor HDR Dataset~\cite{gardner2017learning} is used for training the \baseline model. To generate the input image from the panorama, we extract \num{10} rectified crops from each HDR panoramas in the training set. As in \cite{weber2022editable}, the resulting images are converted to LDR by re-exposing to make the median intensity equal to \num{0.45}, clipping to \num{1}, and applying a $\gamma=\nicefrac{1}{2.2}$ tonemapping. The same exposure factor is subsequently applied to the target HDR panorama to ensure consistency. 

In addition, a subset of the Zillow Indoor Dataset~\cite{cruz2021zillow} is used for training, which contains a total of \num{67448} LDR indoor panoramas of \num{1575} unfurnished residences. This dataset contains LDR data, so no high dynamic range lighting values are available. However, this dataset is useful to provide additional visual examples of indoor scenes so the network can learn to generate more realistic outputs. We use a subset of the same size as the Laval dataset. 

\paragraph{Loss functions} 
A combination of GAN, feature matching, perceptual losses are employed~\cite{wang2018high}. The same default weights as in \cite{wang2018high} are used in training. Since the Zillow dataset does not contain HDR, we ensure we do not penalize the network for generating HDR values by clipping its output $L$ to the $[0,1]$ interval before computing the losses. That way, any produced HDR value will not be considered in the losses. Since $L$ is of equirectangular projection, the feature matching loss is spatially weighted by the solid angle at each pixel. The discriminator employs minibatch standard deviation~\cite{karras2017progressive} to minimize mode collapse. 

Since we are interested in studying the impact of color on the task of virtual object insertion for augmented reality, we also employ a render loss. To compute this loss, we render a virtual scene with the predicted lighting and compare it with the same virtual scene rendered with the ground truth. Since the scene is static, we pre-compute a light transport matrix $T$:
\begin{equation}
\ell_\text{render} = ||TL - TL^*||_1 \,.
\end{equation} 
Here, $T$ is the transport matrix for a lambertian scene (without interreflections) composed of a flat ground plane with 9 spheres arranged in a grid fashion, observed from above as in \cite{weber2022editable}.

\begin{figure}[t]
\centering
\footnotesize
\setlength{\tabcolsep}{0pt}
\begin{tabular}{cc}
\includegraphics[width=0.5\linewidth, trim=0.2cm 0 0.1cm 0.1cm, clip]{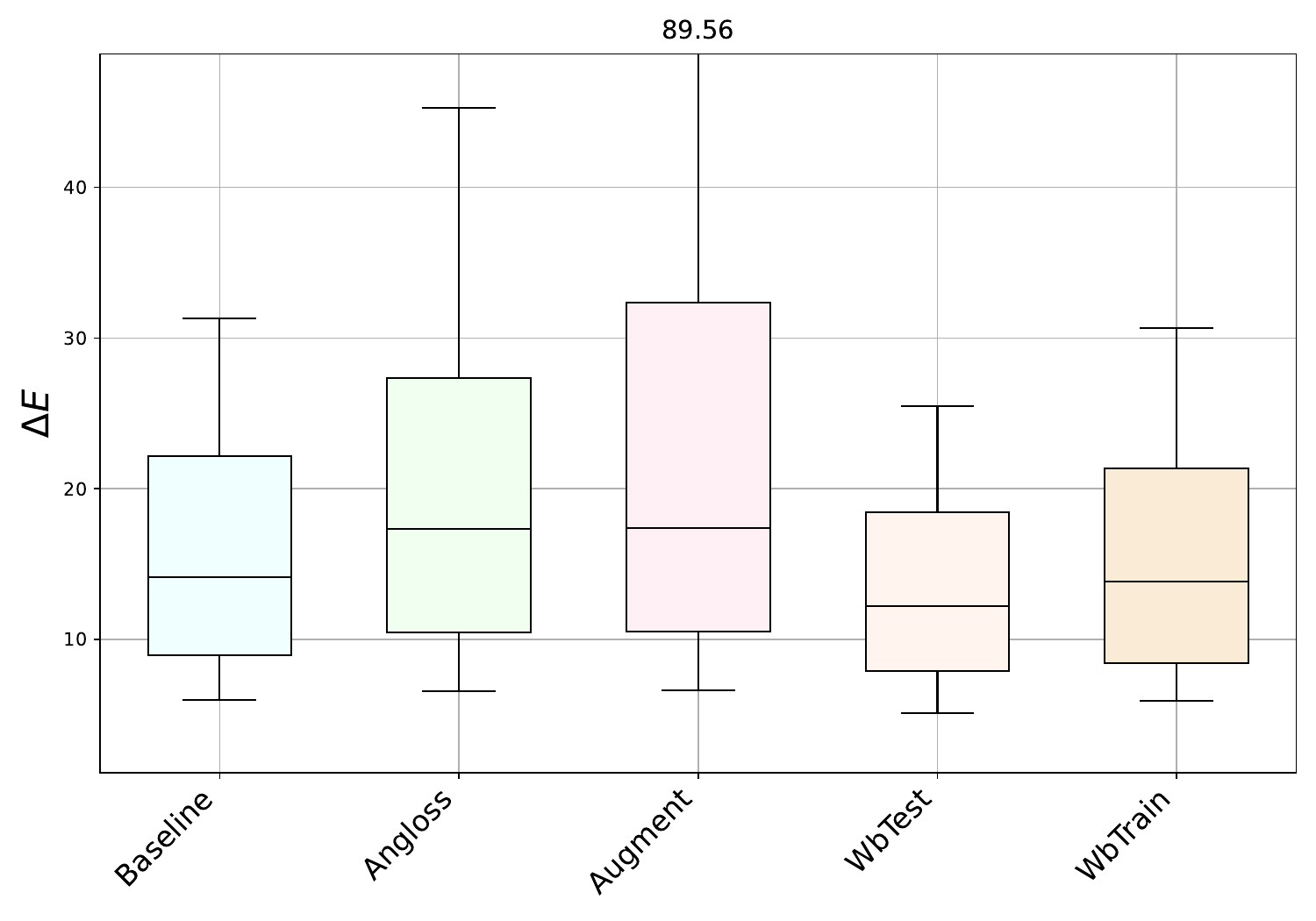} &
\includegraphics[width=0.5\linewidth, trim=0.2cm 0 0.1cm 0.1cm, clip]{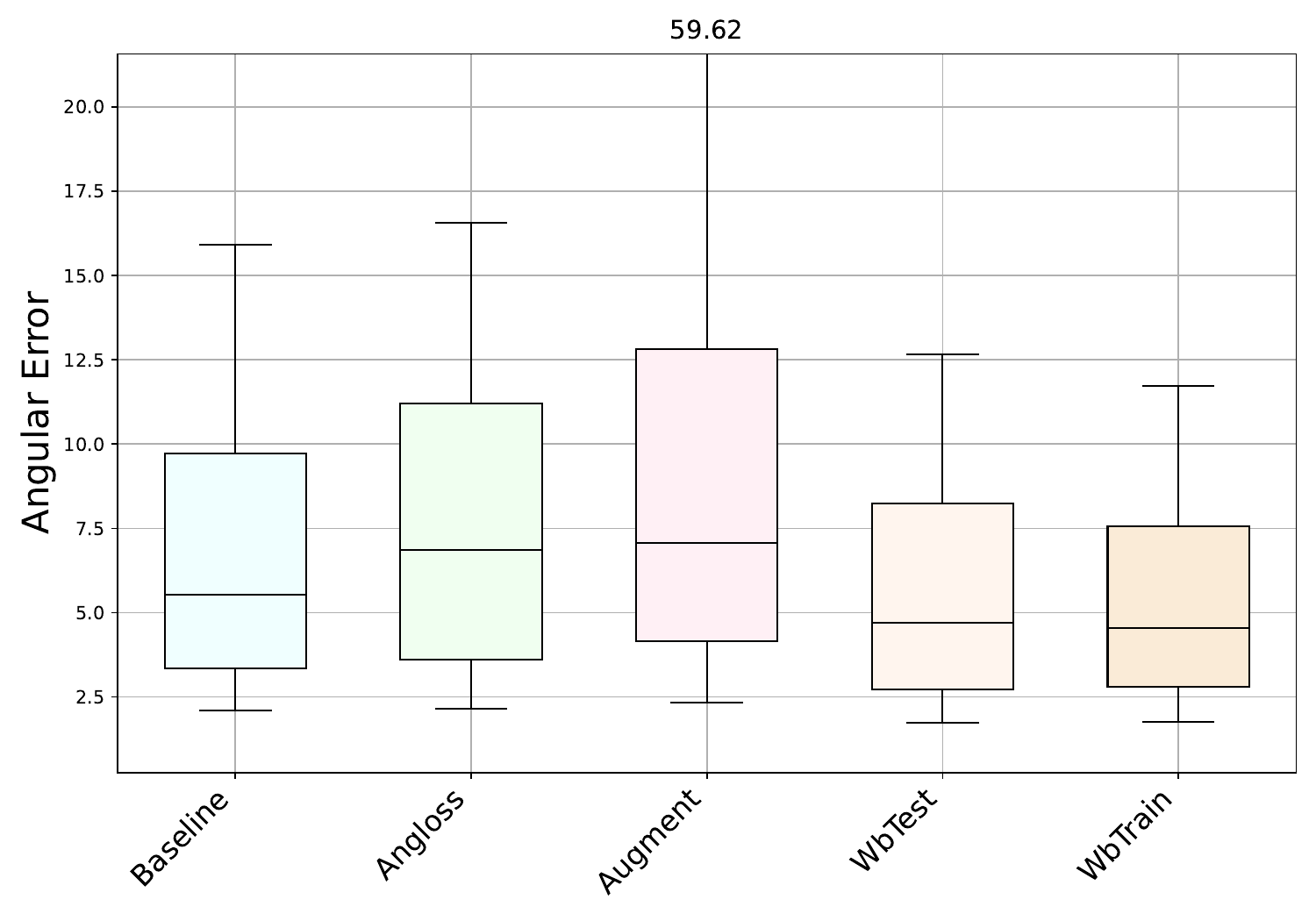} \\ 
(a) $\Delta E$ & 
(b) RGB angular error \\
% \includegraphics[width=0.5\linewidth, trim=0.2cm 0 0.1cm 0.1cm, clip]{figures/graphs/boxplot/rmse_boxplot.pdf} &
% \includegraphics[width=0.5\linewidth, trim=0.2cm 0 0.1cm 0.1cm, clip]{figures/graphs/boxplot/si_rmse_boxplot.pdf} \\
% (c) RSME & 
% (d) si-RMSE \\
\end{tabular}
\caption{Distribution of errors on our test dataset for the (a) $\Delta E$ and (b) RGB angular error metrics for the different color adaptation strategies.}
\label{fig:quant-boxplots}
\end{figure}

\begin{figure}[t]
\centering
\footnotesize
\setlength{\tabcolsep}{0pt}
\begin{tabular}{cccc}
\multicolumn{2}{c}{
	\includegraphics[width=0.75\linewidth, trim=0.2cm 1.5cm 0.2cm 1.5cm, clip]{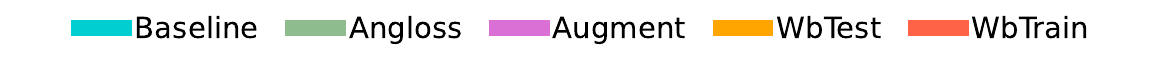}
} \\
\includegraphics[height=4.4cm, trim=1cm 0 1.5cm 1.75cm, clip]{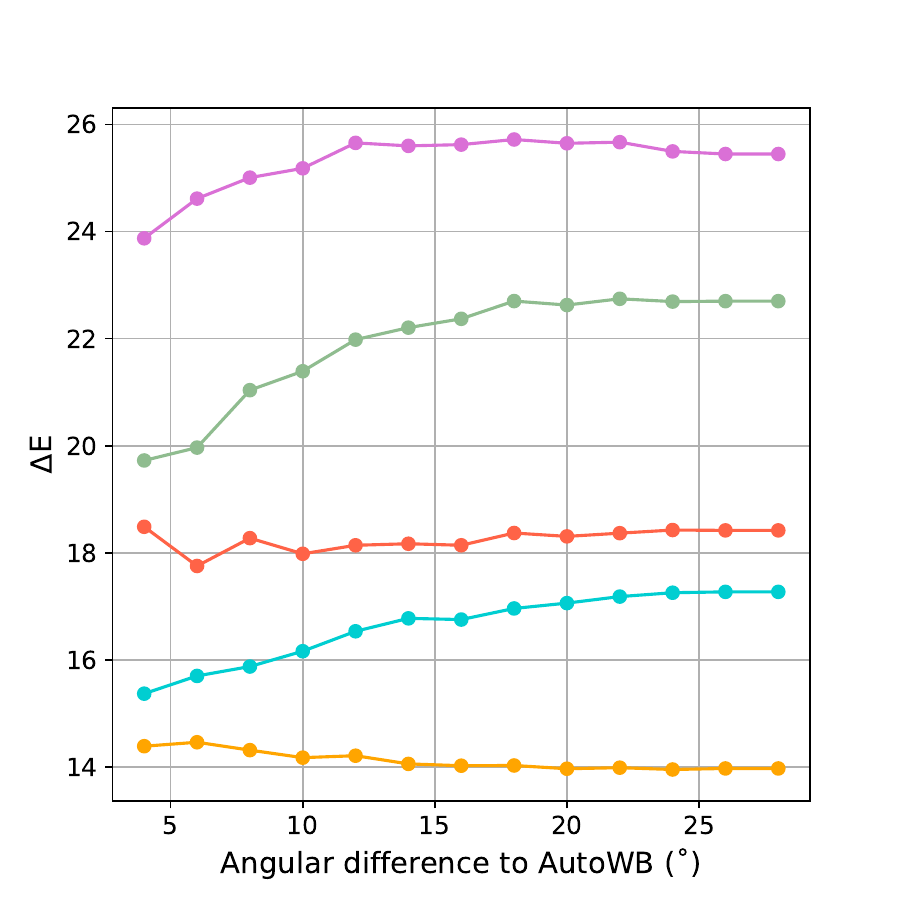} & 
\includegraphics[height=4.4cm, trim=1cm 0 1.5cm 1.75cm, clip]{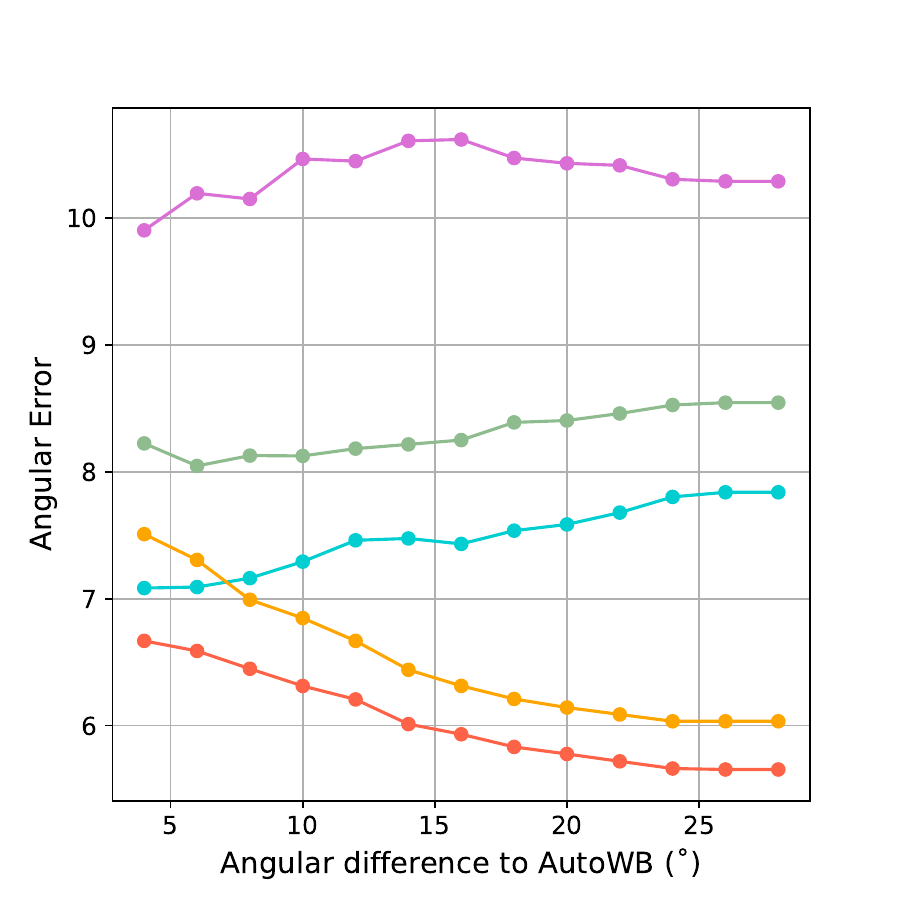} \\
(a) $\Delta E$ & 
(b) RGB angular error \\
% \includegraphics[height=4.4cm, trim=0.5cm 0 1.5cm 1.75cm, clip]{figures/graphs/curves/RMSE_metrics.pdf} & 
% \includegraphics[height=4.4cm, trim=0.25cm 0 1.5cm 1.75cm, clip]{figures/graphs/curves/si-RMSE_metrics.pdf} \\
% (d) RSME & 
% (e) si-RMSE \\
\end{tabular}
\caption{Error metric as a function of the angular distance to the reference automatic white balance (``AutoWB'') setting onboard the camera in the test dataset.}
\label{fig:quant-curves}
\end{figure}

\begin{figure}[t]
\centering
\footnotesize
\setlength{\tabcolsep}{0pt}
\begin{tabular}{cc}
\includegraphics[width=0.5\linewidth, trim=0.2cm 0 0.1cm 0.1cm, clip]{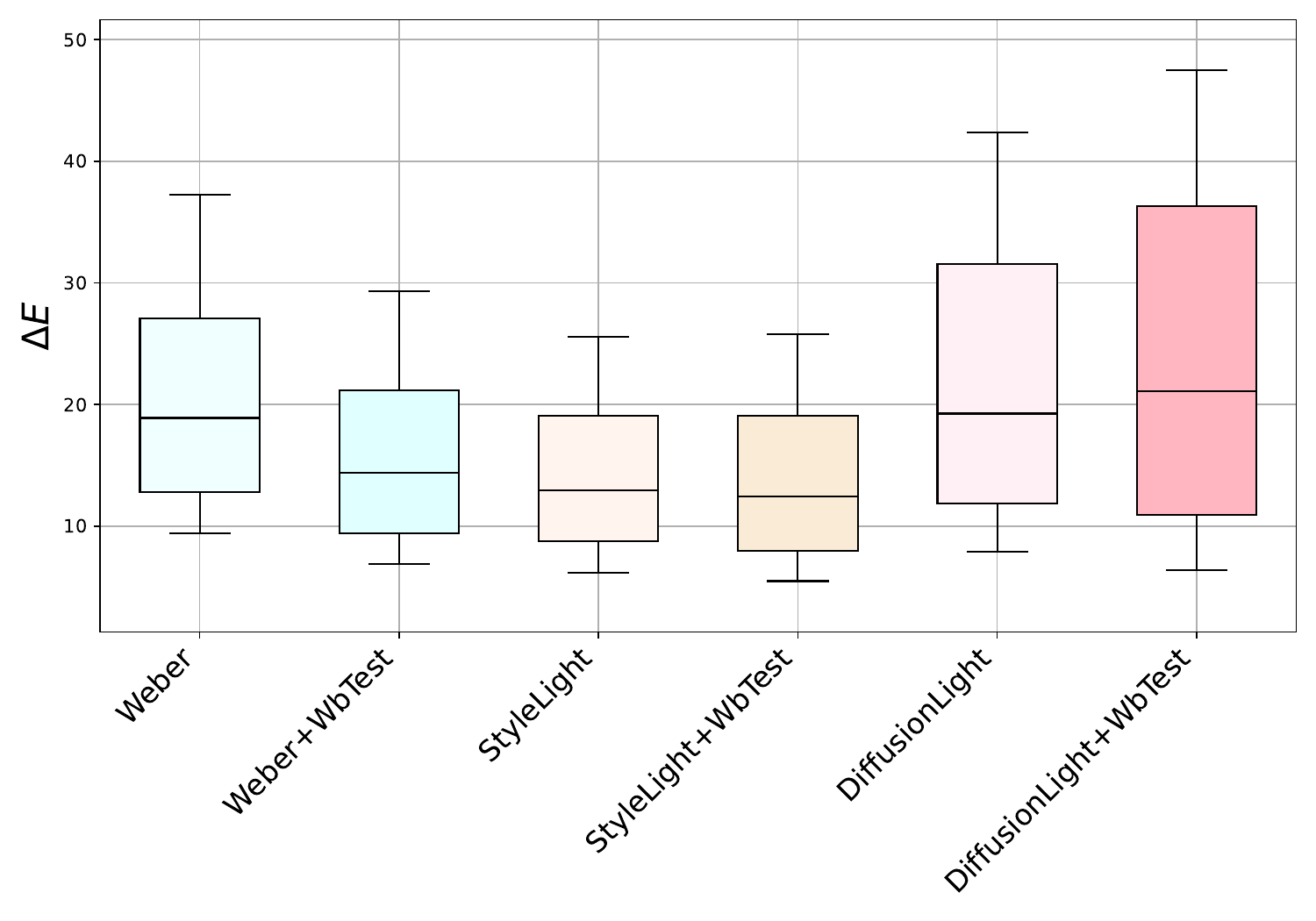} &
\includegraphics[width=0.5\linewidth, trim=0.2cm 0 0.1cm 0.1cm, clip]{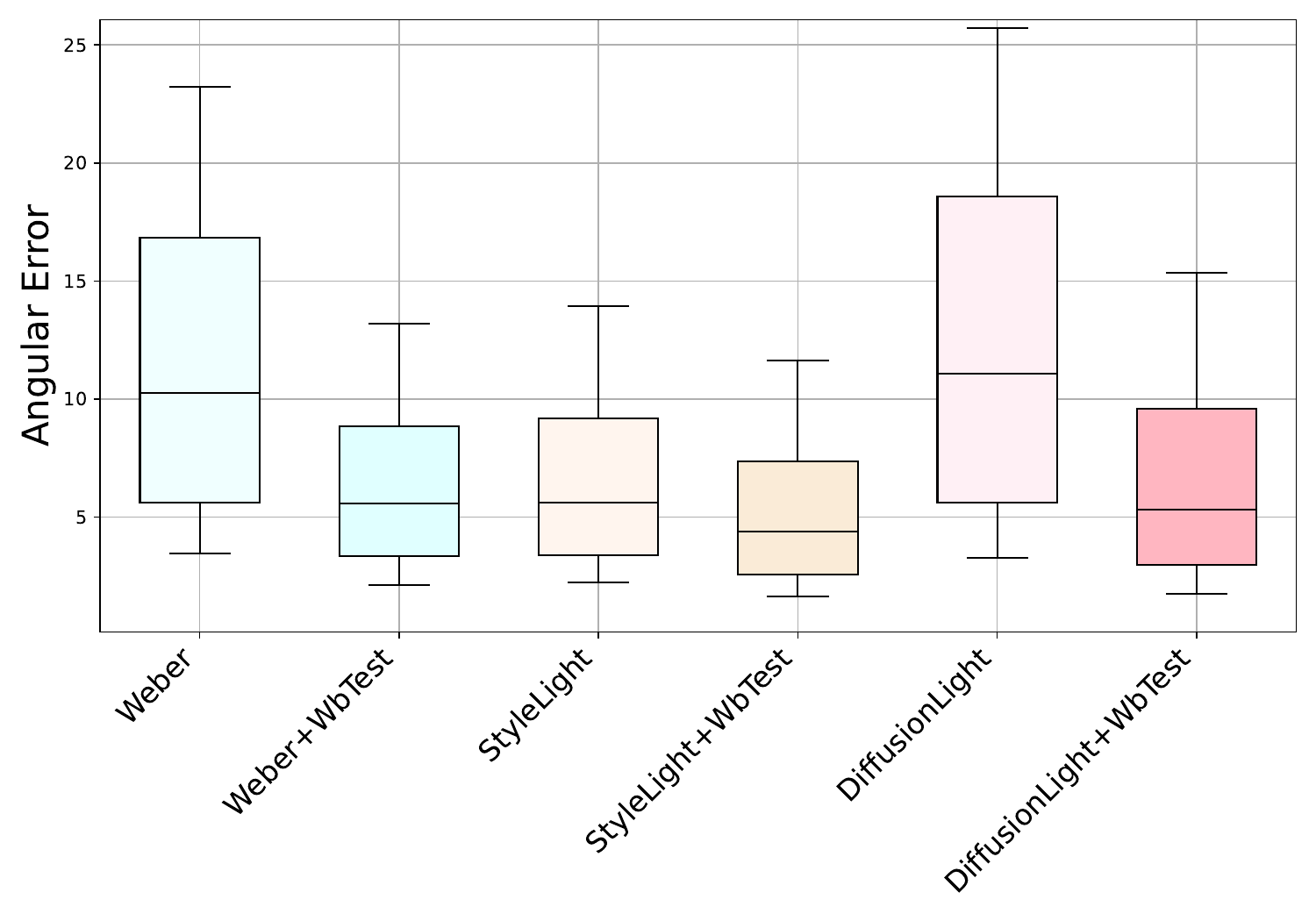} \\ 
(a) $\Delta E$ & 
(b) RGB angular error \\
% \includegraphics[width=0.5\linewidth, trim=0.2cm 0 0.1cm 0.1cm, clip]{figures/graphs_existing/boxplot/rmse_boxplot.pdf} &
% \includegraphics[width=0.5\linewidth, trim=0.2cm 0 0.1cm 0.1cm, clip]{figures/graphs_existing/boxplot/si_rmse_boxplot.pdf} \\
% (d) RSME & 
% (e) si-RMSE \\
\end{tabular}
\caption{Distribution of errors on our test dataset for Weber~\etal~\cite{weber2022editable}, StyleLight~\cite{wang2022stylelight}, and DiffusionLight~\cite{phongthawee2024diffusionlight}. The figure shows the (a) $\Delta E$ and (b) RGB angular error metrics and for the \strategywb strategy.}
\label{fig:quant-boxplots-others}
\end{figure}

% \begin{figure}[t]
% \centering
% \footnotesize
% \setlength{\tabcolsep}{0pt}
% \begin{tabular}{cccc}
% \multicolumn{2}{c}{
% 	\includegraphics[width=0.75\linewidth, trim=0.2cm 1.5cm 0.2cm 1.5cm, clip]{figures/graphs_existing/curves/legend.pdf}
% } \\
% \includegraphics[height=4.4cm, trim=1cm 0 1.5cm 1.75cm, clip]{figures/graphs_existing/curves/ΔE_metrics.pdf} & 
% \includegraphics[height=4.4cm, trim=1cm 0 1.5cm 1.75cm, clip]{figures/graphs_existing/curves/Angular Error_metrics.pdf} \\
% (a) $\Delta E$ & 
% (b) RGB angular error \\
% % \includegraphics[height=4.4cm, trim=0.5cm 0 1.5cm 1.75cm, clip]{figures/graphs_existing/curves/RMSE_metrics.pdf} & 
% % \includegraphics[height=4.4cm, trim=0.5cm 0 1.5cm 1.75cm, clip]{figures/graphs_existing/curves/si-RMSE_metrics.pdf} \\
% % (d) RSME & 
% % (e) si-RMSE \\
% \end{tabular}
% \caption{Error metric as a function of the angular distance to the reference automatic white balance (``AutoWB'') setting onboard the camera in the test dataset, for the methods from Weber~\etal~\cite{weber2022editable} and StyleLight~\cite{wang2022stylelight}.}
% \label{fig:quant-curves-others}
% \end{figure}

\subsection{Evaluation methodology}

We retrain several versions of the baseline network by employing the proposed strategies, and evaluate the trained networks on all the images of our test dataset. We reiterate that the images in the test dataset do not overlap the training set: both were captured with different cameras. To measure performance and since we are interested in the use of lighting estimation for virtual object insertion, we first render a virtual scene with the estimated lighting $L$, and compare it with a rendering of the same virtual scene with the ground truth lighting $L^*$. The virtual scene follows a setup similar to \cite{weber2022editable}, featuring 9 grid-aligned diffuse spheres above a plane shadow catcher, viewed from a bottom-facing camera. Then, the $\Delta E$~\cite{deltaE} and RGB angular error~\cite{Angularerror} metrics are computed as they are color-specific. %, but also measure the others to ensure that they are not negatively affected by our color adaptation strategies. Please refer to the supplement for the other metrics.

% Finally, we also report the FID~\cite{heusel2017gans} for each model, as a proxy for the realism of the estimated lighting maps. Ideally, we would also like the FID to remain as constant as possible with the use of the color-specific strategies.  
% \jf{Do we keep the FID in there? }

\subsection{Experimental results}

We begin by presenting the distribution of errors obtained by applying each of the strategies on our test dataset. The results are shown in \cref{fig:quant-boxplots}. Compared to \baseline, we first observe that several strategies actually make things \emph{worse} in terms of color. Indeed, \strategyangloss and \strategyaug both result in a worsening of the overall color metrics. In contrast, \strategywb and \strategywbtrain improve the color accuracy of lighting estimation. We observe that the best results across all metrics are obtained with \strategywb, which indicates that including a white balance correction step at test time only works better than doing so both at training and test time. This strategy is promising: it is both simpler (as it requires no training) and achieves an improvement in performance that is greater than any of the other strategies tried. We also observe that the worse-case results (shown by the top ticks in the graphs) is significantly improved using \strategywb.

\begin{figure*}
\centering
\scriptsize
% \newlength{\mywidth}
\setlength{\mywidth}{0.14\linewidth}
\setlength{\tabcolsep}{0.5pt}
\def\basepath{figures/qual/results_panos_vis}
\def\first{AtriumElevatorDay/R0015322_Incandescent1_0}
\def\second{VND3262_BrownArea/R0016525_Shade_0}
\def\fifth{VND3400_GreenArmchairs/R0016565_AutoWB_1}
\def\sixth{VND3122_WestChairs/R0016289_Shade_1}

\def\baselinepath{\basepath/baseline+batchstd2+renderloss}
\def\anglosspath{\basepath/baseline+batchstd2+renderloss+angloss}
\def\augmentpath{\basepath/baseline+batchstd2+renderloss+augdata}
\def\wbtestpath{\basepath/baseline+batchstd2+renderloss_inputwb}
\def\wbtrainpath{\basepath/baseline+batchstd2+autowb+renderloss}
\def\gtpath{\basepath/gt}

\begin{tabular}{ccccccc}
% \includegraphics[width=\mywidth]{\baselinepath/\second_input.jpg} &
% \includegraphics[width=\mywidth]{\baselinepath/\second_predict.jpg} & 
% \includegraphics[width=\mywidth]{\anglosspath/\second_predict.jpg} & 
% \includegraphics[width=\mywidth]{\augmentpath/\second_predict.jpg} & 
% \includegraphics[width=\mywidth]{\wbtestpath/\second_predict.jpg} &
% \includegraphics[width=\mywidth]{\wbtrainpath/\second_predict.jpg} \\

%  &
% \includegraphics[width=\mywidth]{\baselinepath/\second_pred_render.jpg} & 
% \includegraphics[width=\mywidth]{\anglosspath/\second_pred_render.jpg} & 
% \includegraphics[width=\mywidth]{\augmentpath/\second_pred_render.jpg} & 
% \includegraphics[width=\mywidth]{\wbtestpath/\second_pred_render.jpg} &
% \includegraphics[width=\mywidth]{\wbtrainpath/\second_pred_render.jpg} \\

% \includegraphics[width=\mywidth]{\baselinepath/\fifth_input.jpg} &
% \includegraphics[width=\mywidth]{\baselinepath/\fifth_predict.jpg} & 
% \includegraphics[width=\mywidth]{\anglosspath/\fifth_predict.jpg} & 
% \includegraphics[width=\mywidth]{\augmentpath/\fifth_predict.jpg} & 
% \includegraphics[width=\mywidth]{\wbtestpath/\fifth_predict.jpg} &
% \includegraphics[width=\mywidth]{\wbtrainpath/\fifth_predict.jpg} \\
%  &
% \includegraphics[width=\mywidth]{\baselinepath/\fifth_pred_render.jpg} & 
% \includegraphics[width=\mywidth]{\anglosspath/\fifth_pred_render.jpg} & 
% \includegraphics[width=\mywidth]{\augmentpath/\fifth_pred_render.jpg} & 
% \includegraphics[width=\mywidth]{\wbtestpath/\fifth_pred_render.jpg} &
% \includegraphics[width=\mywidth]{\wbtrainpath/\fifth_pred_render.jpg} \\

%
\includegraphics[width=\mywidth]{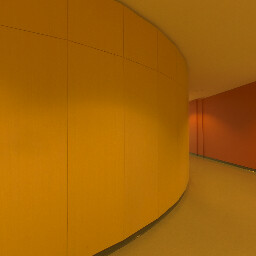} &
\includegraphics[width=\mywidth]{\baselinepath/\sixth_predict.jpg} & 
\includegraphics[width=\mywidth]{\anglosspath/\sixth_predict.jpg} & 
\includegraphics[width=\mywidth]{\augmentpath/\sixth_predict.jpg} & 
\includegraphics[width=\mywidth]{\wbtestpath/\sixth_predict.jpg} &
\includegraphics[width=\mywidth]{\wbtrainpath/\sixth_predict.jpg} &
\includegraphics[width=\mywidth]{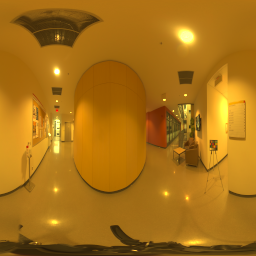} \\*[-2pt]
 &
\includegraphics[width=\mywidth]{\baselinepath/\sixth_pred_render.jpg} & 
\includegraphics[width=\mywidth]{\anglosspath/\sixth_pred_render.jpg} & 
\includegraphics[width=\mywidth]{\augmentpath/\sixth_pred_render.jpg} & 
\includegraphics[width=\mywidth]{\wbtestpath/\sixth_pred_render.jpg} &
\includegraphics[width=\mywidth]{\wbtrainpath/\sixth_pred_render.jpg} &
\includegraphics[width=\mywidth]{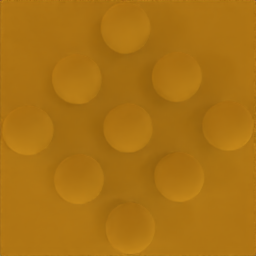} \\

Input & \baseline & \strategyangloss & \strategyaug & \strategywb & \strategywbtrain & GT \\

\end{tabular}
\caption{Qualitative results comparing the estimated HDR environment maps with the various strategies tested. We show the input image (left), the predicted environment maps for each strategy (first row), and a virtual scene rendered with each predicted environment map (bottom row). ``GT'' (right) is the ground truth.}
\label{fig:results-qual}

\end{figure*}

In addition to the statistics computed over the entire test set repored in \cref{fig:quant-boxplots}, we also wish to evaluate how well each strategy work on images when their white balance setting differs from the default auto white balance setting (AWB). For this, recall that for each scene in the test set, we captured images under different white balance settings, including AWB. In \cref{fig:quant-curves}, we thus plot each metric as a function of the RGB angular difference (in degrees) between each test scene and its corresponding AWB image. Here, observe how all metrics progressively become worse as the angular difference increases for \baseline, \strategyangloss and \strategyaug. In contrast, more stable performance is observed for both \strategywb and \strategywbtrain, demonstrating once more that explicitly including white balance correction indeed makes light estimation more robust to color errors. As before, \strategywb, which applies the correction only at test time, is the most stable of all strategies. Qualitative examples are shown in \cref{fig:results-qual} to visually demonstrate the impact of each strategy.

% Present quantitative and qualitative results here. See \cref{fig:quantitative-results} for quantitative results.

% \begin{table}
% \centering
% \caption{FID for each model. }
% \label{tab:fid}
% \begin{tabular}{lc}
% \toprule
% Method & FID \\
% \midrule
% \baseline 			& 165.0311366\\
% \midrule
% \strategyangloss 	& 109.038851 \\
% \strategyaug 		& 161.437650 \\
% \strategywb 		& 170.048330 \\
% \strategywbtrain 	& 97.600357 \\
% \bottomrule
% \end{tabular}
% \end{table}

\subsection{Other lighting estimation baselines}
\label{sec:other-baselines}

\begin{figure*}
\centering

\footnotesize
\setlength{\mywidth}{0.24\linewidth}
\setlength{\tabcolsep}{1pt}
\def\basepath{figures/qual/results_panos_vis}
\def\first{AtriumElevatorDay/R0015322_Incandescent1_0}
\def\second{VND3262_BrownArea/R0016525_Shade_0}
\def\fifth{VND3400_GreenArmchairs/R0016565_AutoWB_1}
\def\sixth{VND3122_WestChairs/R0016289_Shade_1}

\def\baselinepath{\basepath/baseline+batchstd2+renderloss}
\def\anglosspath{\basepath/baseline+batchstd2+renderloss+angloss}
\def\augmentpath{\basepath/baseline+batchstd2+renderloss+augdata}
\def\wbtestpath{\basepath/baseline+batchstd2+renderloss_inputwb}
\def\wbtrainpath{\basepath/baseline+batchstd2+autowb+renderloss}

\begin{tabular}{cccc}

\multirow{2}{*}[14ex]{\includegraphics[width=\mywidth]{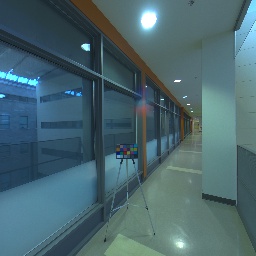}} &
\includegraphics[width=\mywidth]{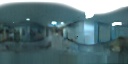} &
\includegraphics[width=\mywidth]{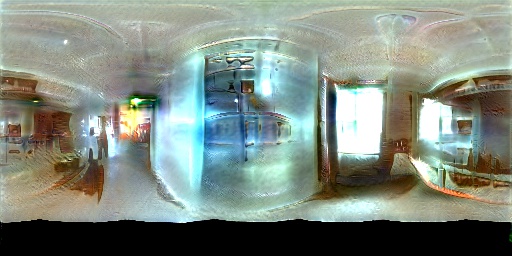} & 
\includegraphics[width=\mywidth]{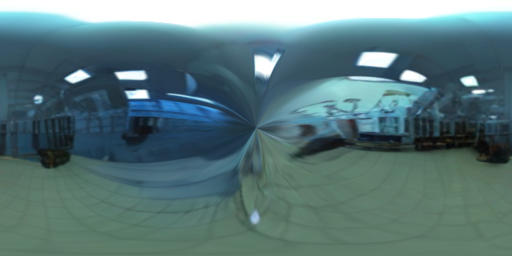} \\
& 
\includegraphics[width=\mywidth]{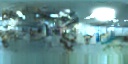} &
\includegraphics[width=\mywidth]{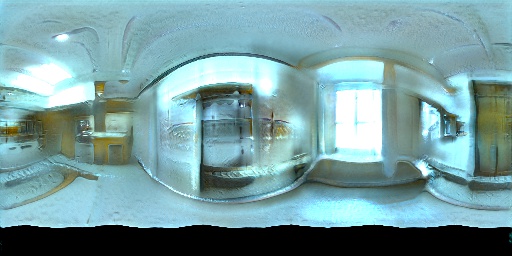} &
\includegraphics[width=\mywidth]{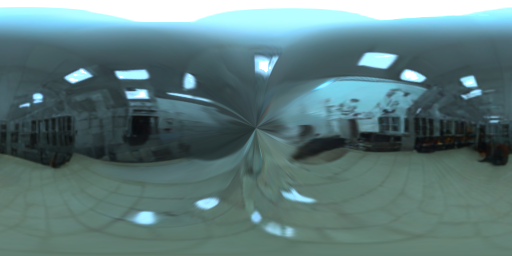} \\*[0.25em]

% \multirow{2}{*}[5.5ex]{\includegraphics[width=\mywidth]{figures/qual_existing/R0017142_Incandescent1_0_input.jpg}} &
% \includegraphics[width=\mywidth]{figures/qual_existing/R0017142_Incandescent1_0_predict_henrique.jpg} &
% \includegraphics[width=\mywidth]{figures/qual_existing/R0017142_Incandescent1_0_predict_stylelight.jpg} \\
% & 
% \includegraphics[width=\mywidth]{figures/qual_existing/R0017142_Incandescent1_0_predict_henrique_inputwb.jpg} &
% \includegraphics[width=\mywidth]{figures/qual_existing/R0017142_Incandescent1_0_predict_stylelight_inputwb.jpg} \\*[0.25em]

% \multirow{2}{*}[5.5ex]{\includegraphics[width=\mywidth]{figures/qual_existing/R0016098_Incandescent1_2_input.jpg}} &
% \includegraphics[width=\mywidth]{figures/qual_existing/R0016098_Incandescent1_2_predict_henrique.jpg} &
% \includegraphics[width=\mywidth]{figures/qual_existing/R0016098_Incandescent1_2_predict_stylelight.jpg} \\
% & 
% \includegraphics[width=\mywidth]{figures/qual_existing/R0016098_Incandescent1_2_predict_henrique_inputwb.jpg} &
% \includegraphics[width=\mywidth]{figures/qual_existing/R0016098_Incandescent1_2_predict_stylelight_inputwb.jpg} \\*[0.25em]

\multirow{2}{*}[14ex]{\includegraphics[width=\mywidth]{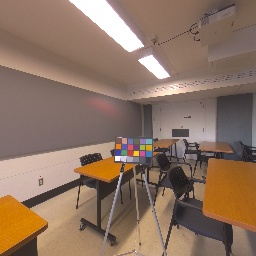}} &
\includegraphics[width=\mywidth]{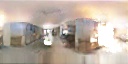} &
\includegraphics[width=\mywidth]{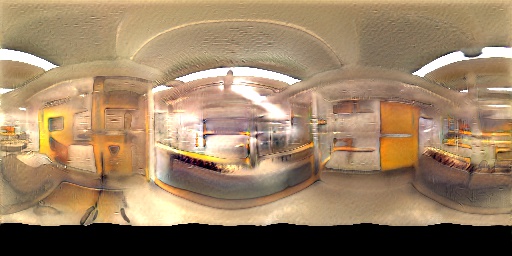} &
\includegraphics[width=\mywidth]{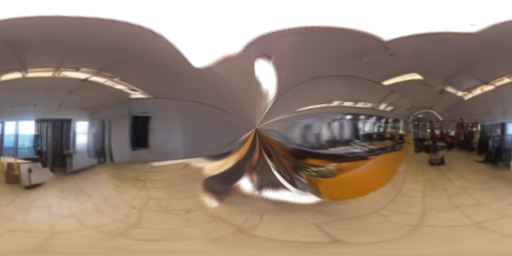} \\
& 
\includegraphics[width=\mywidth]{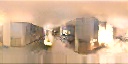} &
\includegraphics[width=\mywidth]{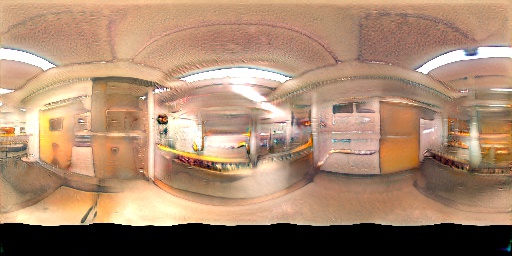} &
\includegraphics[width=\mywidth]{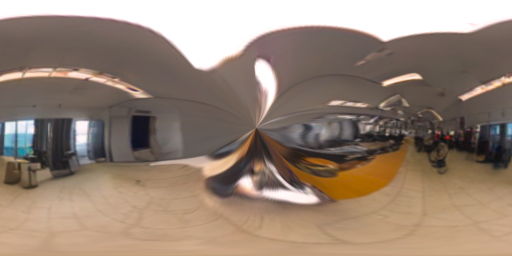} \\*[0.25em]

\multirow{2}{*}[14ex]{\includegraphics[width=\mywidth]{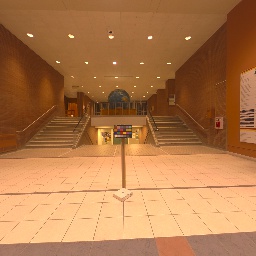}} &
\includegraphics[width=\mywidth]{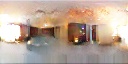} &
\includegraphics[width=\mywidth]{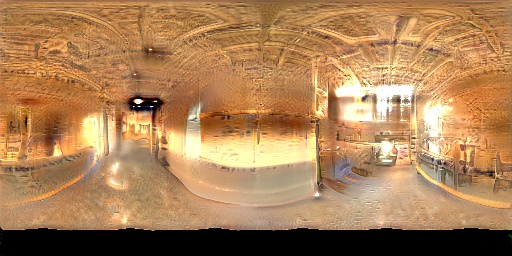} &
\includegraphics[width=\mywidth]{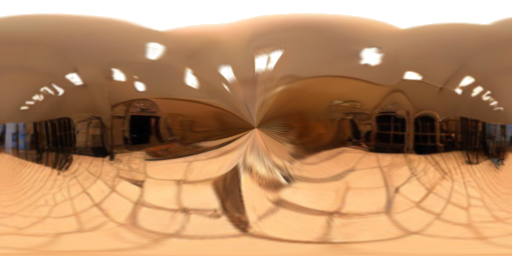} \\
& 
\includegraphics[width=\mywidth]{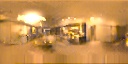} &
\includegraphics[width=\mywidth]{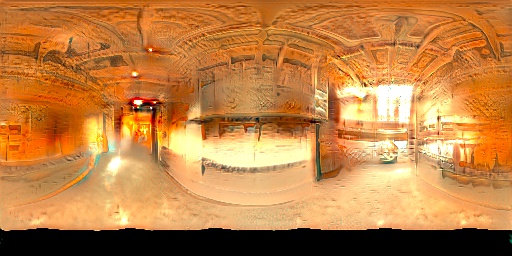} &
\includegraphics[width=\mywidth]{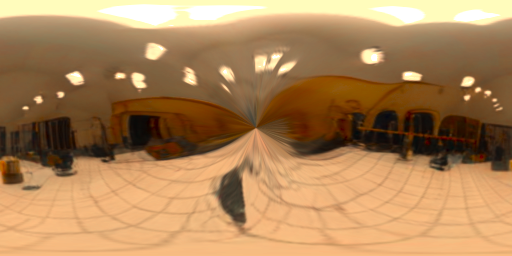} \\
Input image & Weber & StyleLight & DiffusionLight \\
\end{tabular}

\footnotesize
\setlength{\mywidth}{0.15\linewidth}
\setlength{\tabcolsep}{1pt}

\caption{Qualitative results comparing the estimated HDR environment maps when applying the \strategywb strategy to the techniques from Weber~\etal~\cite{weber2022editable}, StyleLight~\cite{wang2022stylelight}, and DiffusionLight~\cite{phongthawee2024diffusionlight}. Original (top) and color-corrected (bottom) predictions are shown.}
\label{fig:results-qual-others}
\end{figure*}

It was established that the \strategywb strategy resulted in improved color accuracy when estimating lighting, when evaluated on our \baseline model. Here, we evaluate whether these findings can be corroborated with other lighting estimation models. Specifically, we experiment with Weber~\etal~\cite{weber2022editable}, StyleLight~\cite{wang2022stylelight} and DiffusionLight~\cite{phongthawee2024diffusionlight}. As observed in \cref{fig:quant-boxplots-others}, we observe that \strategywb improves color robustness quite significantly, especially visible in the angular error metric.
 % for Weber. The difference is less striking for StyleLight, as this approach seems to be more robust already. Importantly, the performance is not negatively affected by the color balancing strategy. 
This is corroborated visually in \cref{fig:results-qual-others}, which illustrates qualitative examples applying \strategywb to both techniques. While the difference is less noticeable for StyleLight and DiffusionLight, the color balance can still be improved.

%!TEX root = ../main_cic.tex

% \input{figures/fig_qual_others2}
%!TEX root = ../main_cic.tex

% \subsection{Failure cases and limitations}

% \jf{remove FID}

%!TEX root = ../main_cic.tex
\section{Discussion}
\label{sec:discussion}

This paper examines the color robustness of lighting estimation algorithms. Rather than proposing a new method, our focus is on developing simple yet effective strategies to enhance the color accuracy of \emph{existing} algorithms. To this end, we introduce a new dataset of indoor HDR lighting environments and simulate varying color conditions by applying different on-camera white balance presets. We establish a baseline lighting estimation model and use it to evaluate several color adaptation strategies: modifying the loss function (\strategyangloss), applying data augmentation (\strategyaug), and incorporating a white balance network either at test time (\strategywb) or during training (\strategywbtrain). Our experiments demonstrate that \strategywb offers a particularly robust way to improve a pre-trained lighting estimation network's resilience to variations in illuminant color. Importantly, this approach requires no retraining of the original model. We validate its generality by applying it to two additional lighting estimation methods from the recent literature.

While the paper presents practical strategies and a valuable dataset for exploring color robustness, we also acknowledge some limitations that can inform future work. First, simulating different illuminants through camera white balance settings is not equivalent to capturing scenes illuminated by physically different light sources (\eg, as in \cite{kim2021large}). In some cases, the chosen white balance settings significantly deviate from the scene’s actual illumination, resulting in unrealistic color shifts that would not typically occur with automatic white balance in real-world scenarios. Although our dataset includes such cases intentionally as a way to stress-test models under extreme conditions, future work could improve realism by capturing scenes under varied, physically accurate illuminants. %, ideally using $360^\circ$ imagery to capture the full surrounding lighting.
Second, our baseline model relies on a GAN-based architecture, which introduces its own set of challenges. GANs are well known for training instability and issues such as mode collapse~\cite{lucic2018gans}, phenomena we also observed during our experiments. While we employed techniques to mitigate these issues (\eg, \cite{karras2017progressive}), these problems are specific to GANs and might be avoided altogether by exploring alternative architectures---such as diffusion models---which offer greater stability and are increasingly popular in generative modeling tasks.

% \paragraph*{Limitations} White balance isn't different illuminants. Creates extremes that might not be realistic. 

% Surprisingly, two of these strategies (\strategyangloss and \strategyaug) makes things worse---we hypothesize this might be due to the 

%\section{Acknowledgments} 
%add the acknowledgement section here


%%%%%%%%%%%%%%%%%%%%%%%%%%%%%%%%%%
% Bibliography
%%%%%%%%%%%%%%%%%%%%%%%%%%%%%%%%%%

{
\footnotesize
\bibliographystyle{abbrv}
\bibliography{main}
}


% \begin{biography}
% Please submit a brief biographical sketch of no more than 75 words. 
% Include relevant professional and educational information as shown ;
% in the example below.

% Jane Doe received her BS in physics from the University of Nevada (1977) 
% and her PhD in applied physics from Columbia University (1983). Since 
% then she has worked in the Research and Technology Division at Xerox 
% in Webster, NY. Her work has focused on the development of toner adhesion 
% and transport issues. She is on the Board of  IS\&T and a member of APS 
% and SPIE.
% \end{biography}

\end{document}